%% file: main.tex
\documentclass[10pt,twocolumn,letterpaper]{article}

\usepackage[final]{cvpr}              %

\usepackage{times}
\usepackage{graphicx}
\usepackage{amsmath}
\usepackage{amssymb}

\usepackage{booktabs} %
\usepackage[table]{xcolor}
\usepackage{colortbl}
\usepackage{multirow}
\usepackage[font=footnotesize]{caption}
\usepackage{placeins}

\usepackage{xspace}

\usepackage[pagebackref,breaklinks,colorlinks]{hyperref}

\usepackage[capitalize]{cleveref}
\crefname{section}{Sec.}{Secs.}
\Crefname{section}{Section}{Sections}
\Crefname{table}{Table}{Tables}
\crefname{table}{Tab.}{Tabs.}

\def \etal {\emph{et al.\xspace}}

\def\ournet {PatchConvNet\xspace}
\def\ours {Ours\xspace}

\def \pzo {\phantom{0}} 
\def \dzo {\phantom{00}} 
\def \tzo {\phantom{000}}

\definecolor{Goldenrod}{RGB}{245,245,220}

\begin{document}

\title{Augmenting Convolutional networks with attention-based aggregation}

\author{\begin{minipage}{\linewidth}
\begin{center}
\scalebox{1.0}{\normalsize Hugo Touvron$^{1,2}$ \hspace{0.35cm} Matthieu Cord$^{2}$ \hspace{0.35cm} Alaaeldin El-Nouby$^{1,3}$ \hspace{0.35cm} Piotr Bojanowski$^{1}$ \hspace{0.35cm}}\\[0.1cm]
\scalebox{1.0}{\normalsize Armand Joulin$^{1}$ \hspace{0.35cm} Gabriel Synnaeve$^{1}$ \hspace{0.35cm}   Herv\'e J\'egou$^{1}$}
\\[0.2cm]
\scalebox{1.0}{\normalsize \textmd{$^1$Meta AI\hspace{0.6cm} $^2$Sorbonne University\hspace{0.6cm} $^3$Inria}}
\\[2.0cm]
\end{center}
\end{minipage}
}

\makeatletter
\let\inserttitle\@title
\makeatother
\maketitle

\begin{abstract} 
We show how to augment any convolutional network with an attention-based global map to achieve non-local reasoning. We replace the final average pooling by an attention-based aggregation layer akin to a single transformer block, that weights how the patches are involved in the classification decision. We plug this learned aggregation layer with a simplistic patch-based convolutional network parametrized by 2 parameters (width and depth). In contrast with a pyramidal design, this architecture family maintains the input patch resolution across all the layers. It yields surprisingly competitive trade-offs between accuracy and complexity, in particular in terms of memory consumption, as shown by our experiments on various computer vision tasks: object classification, image segmentation and detection. %
\end{abstract}

\input{introduction}
\input{related}
\input{method}

\input{experiments}

\input{conclusion}

{\small
\bibliographystyle{ieee_fullname}
\bibliography{egbib}
}
\clearpage
\input{appendix}
\end{document}

%% file: introduction.tex
\section{Introduction}

Vision transformers~\cite{dosovitskiy2020image} (ViT) emerge as an alternative to convolutional neural networks (convnets) in computer vision. 
They differ from traditional convnets in many ways, one of which being the patch based processing. Another difference is the aggregation of the image information based on a so-called ``class token''. 
This element correlates with the patches most related to the classification decision. Therefore, the softmax in the self-attention blocks, especially in the last layers, can be used to produce attention maps showing the interaction between the class token and all the patches. 
Such maps have been employed for visualization purposes~\cite{caron2021emerging,dosovitskiy2020image}. It gives some hints on which regions of a given image  are employed by a model to make its decision. 
However the interpretability remains loose: producing these maps involves some fusion of multiple softmax in different different layers and heads.

\begin{figure}[t]
            \vspace{-0.2ex}
          ~~Original ~~~~~~ ViT-S ~~ ``ResNet-50'' ~~~~~ S60 ~~~~~~~~~~ S60$^\dagger$  \\
         \includegraphics[width = 0.19\linewidth]{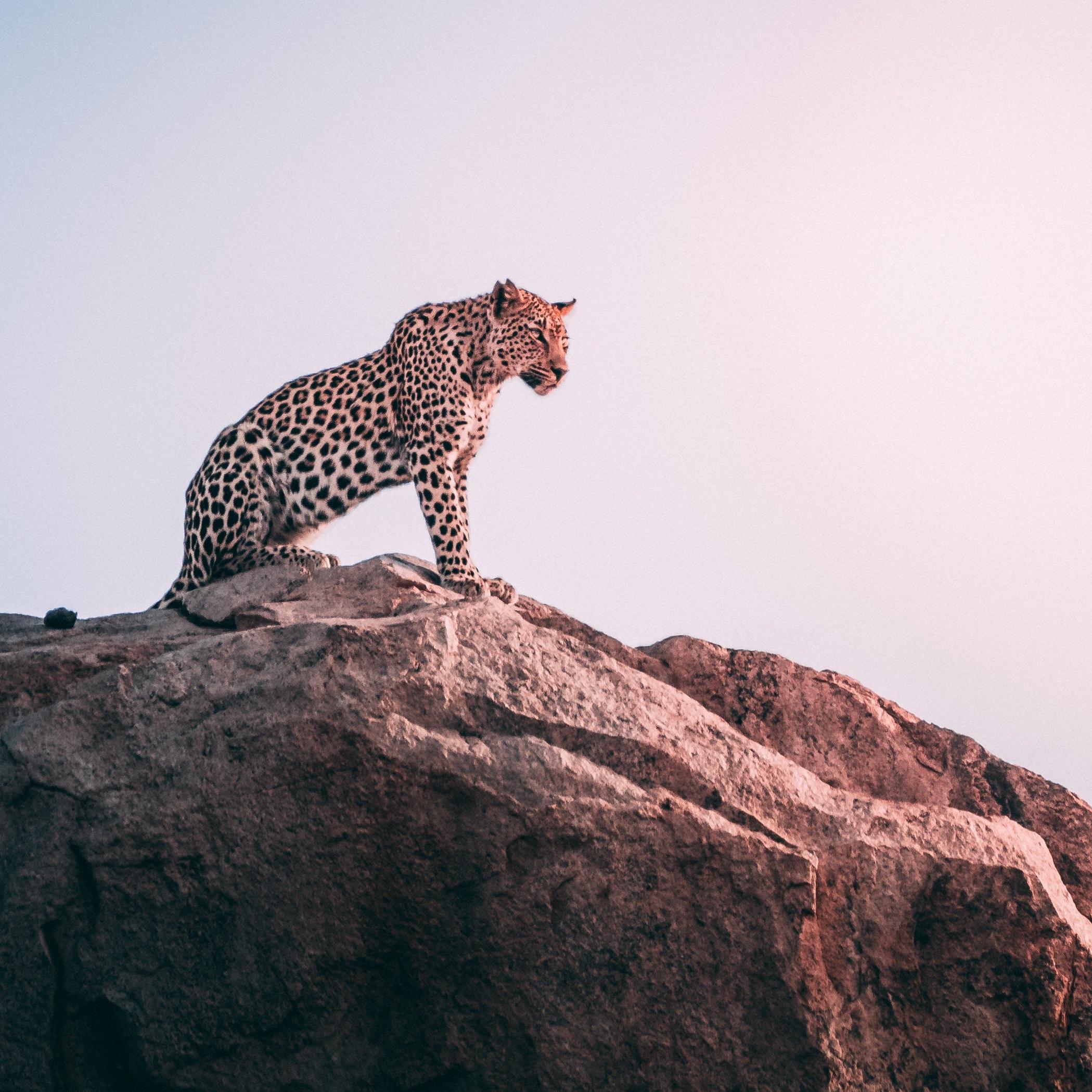} \hfill
         \includegraphics[width = 0.19\linewidth]{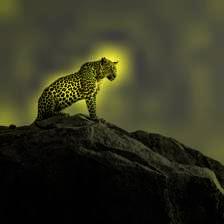} \hfill
          \includegraphics[width = 0.19\linewidth]{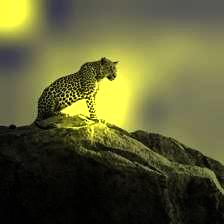} \hfill
         \includegraphics[width = 0.19\linewidth]{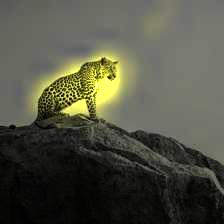} \hfill
         \includegraphics[width = 0.19\linewidth]{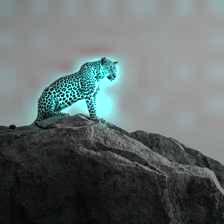}\\
         \includegraphics[width = 0.19\linewidth]{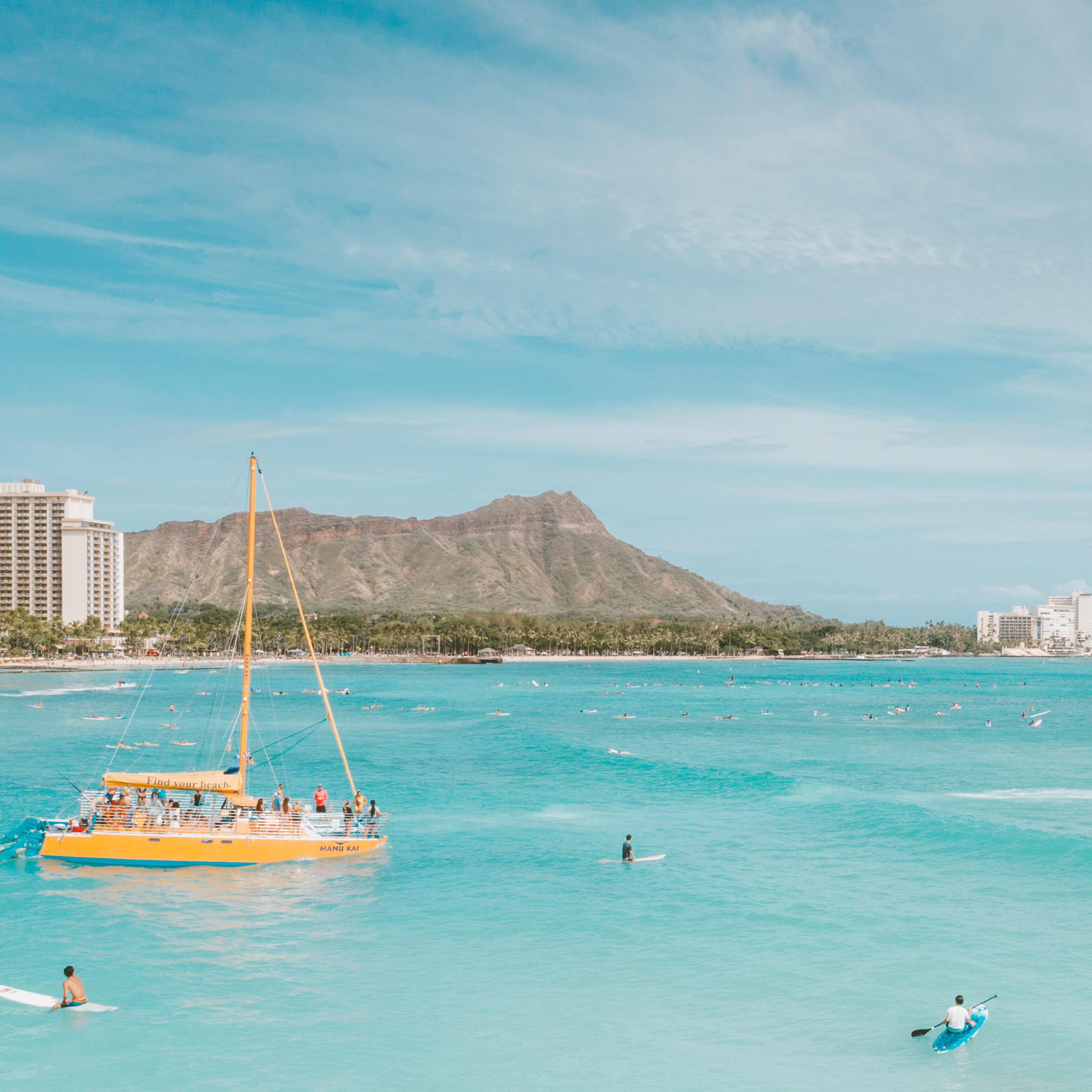} \hfill
         \includegraphics[width = 0.19\linewidth]{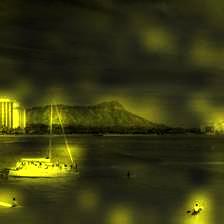} \hfill
        \includegraphics[width = 0.19\linewidth]{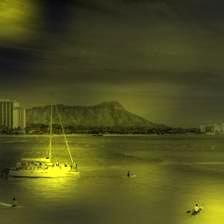} \hfill
         \includegraphics[width = 0.19\linewidth]{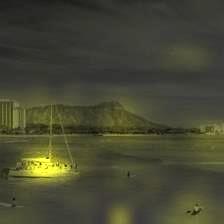} \hfill
         \includegraphics[width = 0.19\linewidth]{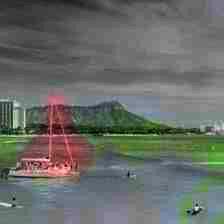}\\
        \includegraphics[width = 0.19\linewidth]{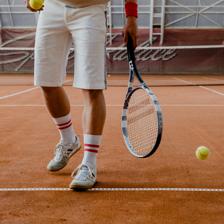} \hfill
         \includegraphics[width = 0.19\linewidth]{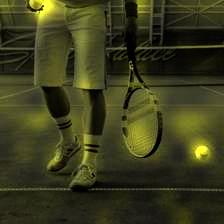} \hfill
        \includegraphics[width = 0.19\linewidth]{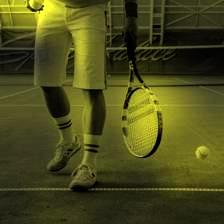} \hfill
         \includegraphics[width = 0.19\linewidth]{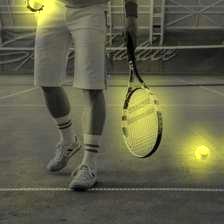} \hfill
         \includegraphics[width = 0.19\linewidth]{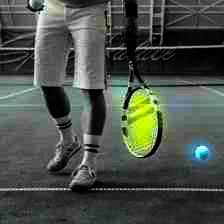}\\
        \includegraphics[width = 0.19\linewidth]{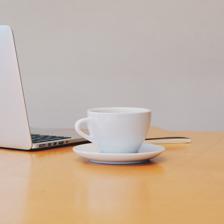} \hfill
         \includegraphics[width = 0.19\linewidth]{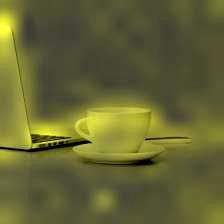} \hfill
        \includegraphics[width = 0.19\linewidth]{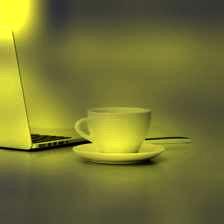} \hfill
         \includegraphics[width = 0.19\linewidth]{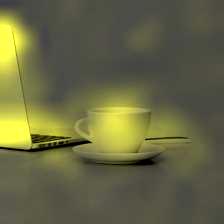} \hfill
         \includegraphics[width = 0.19\linewidth]{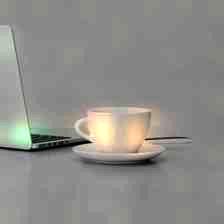}\\
    \vspace{-3.0ex}
    \caption{We augment convolutional neural networks with a learned attention-based aggregation layer. We visualize the attention maps for classification for diverse models. We first extract attention maps from a regular ViT-S~\cite{dosovitskiy2020image,Touvron2020TrainingDI} with Dino-style~\cite{caron2021emerging} vizualizations. Then we consider convnets in which we replace the average pooling by our learned attention-based aggregation layer. Unlike ViT, this layer directly provides the contribution of the patches in the weighted pooling. 
    This is shown for a ``ResNet-50~\cite{He2016ResNet}'', and with our new simple patch-based model (\ournet{}-S60) that we introduce to increase the attention map resolution. 
    We can specialize this attention per class, as shown with S60$\dagger$. 
    \label{fig:attention_maps_intro}}
\end{figure}

\begin{figure}[t]

         \vspace{-0.2ex}
             ~~~~Original~~~~~~~~~~~~~Top-1~~~~~~~~~~~~~~Top-2~~~~~~~~~~~~~~Top-3\\
              \includegraphics[width=0.24\linewidth]{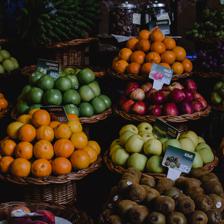} \hfill%
              \includegraphics[width=0.24\linewidth]{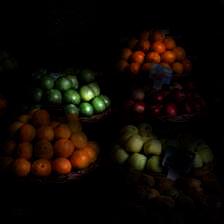} \hfill 
              \includegraphics[width=0.24\linewidth]{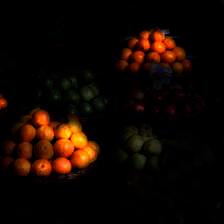} \hfill 
              \includegraphics[width=0.24\linewidth]{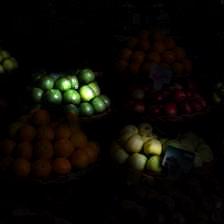} \\[-5pt] %
               {\scriptsize \phantom{0} \hspace{0.215\linewidth} grocery\,store\,(46.7\%) ~~~ orange\,(6.0\%) ~~~~ Granny\,Smith\,(4.9\%)} \\ %
              \includegraphics[width=0.24\linewidth]{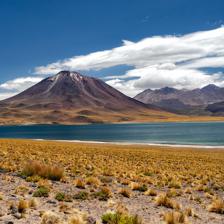} \hfill%
              \includegraphics[width=0.24\linewidth]{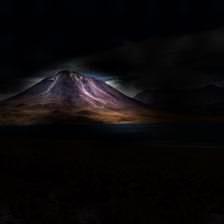} \hfill %
              \includegraphics[width=0.24\linewidth]{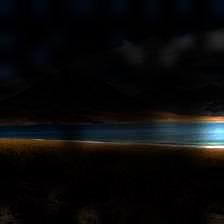} \hfill %
              \includegraphics[width=0.24\linewidth]{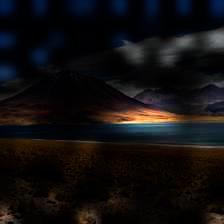}  \\[-5pt] %
             {\scriptsize \phantom{0} \hspace{0.215\linewidth} ~~~volcano\,(86.6\%) ~~~~~~~~~ lakeside\,(0.4\%) ~~~~~~~~ valley\,(0.2\%)}%
              \\
              \includegraphics[width=0.24\linewidth]{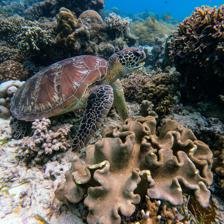} \hfill%
              \includegraphics[width=0.24\linewidth]{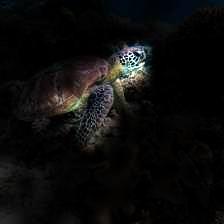} \hfill %
              \includegraphics[width=0.24\linewidth]{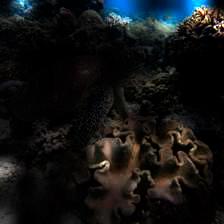} \hfill %
              \includegraphics[width=0.24\linewidth]{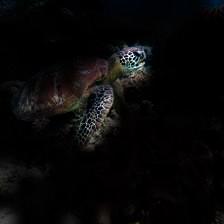}  \\[-5pt] %
             {\scriptsize \phantom{0} \hspace{0.18\linewidth} loggerhead\,turtle\,(66.4\%) ~~coral\,reef\,(15.2\%) ~~leathery\,turtle\,(1.3\%)}%
              \\

        \vspace{-3ex}
    \caption{
    We provide three images for which the attention-based aggregation stage is specialized so as to provide one attention map per class.  We display the attention for the top-3 classes w.r.t. the model prediction. 
    \label{fig:visu_attention}}
\end{figure}

In this paper, we want to provide similar vizualization properties to convnets: %
we augment convnets with an attention map.  %
More precisely we replace the usual average pooling layer by an attention-based layer. 
Indeed, nothing in the convnets design precludes replacing their pooling by attention~\cite{Bello2019AttentionAC}. 
We simplify the design of this attention-based pooling layer such that it explicitly provides the weights of the different patches. %
Compared to ViT, for which the aggregation is performed across multiple layers and heads, our proposal offers a single weight per patch, and therefore a simple way to interpret the attention map: it is the respective contribution of each patch in the weighted sum summarizing the images. 
This treatment allows the model to deal with visual objects separately or jointly: if we use one token for each class instead of a single token, as exemplified in Figures~\ref{fig:attention_maps_intro} and~\ref{fig:visu_attention}, then we obtain an attention weight per patch for each possible class. In our main proposal we mostly focus on the single token case, which is more directly related to the classification decision.

In Figure~\ref{fig:attention_maps_intro}, we show the attention maps extracted from ViT by using a visualization procedure inspired by Caron et al.~\cite{caron2021emerging}. It involves some post-processing as there are multiple layers and heads providing patch weights. 
Then we show a "ResNet-50" augmented by adding our attention-based aggregation layer. Its hierarchical design leads to a low-resolution  attention map with artefacts: We need an architecture producing a higher-resolution feature maps in order to better leverage the proposed attention-based pooling. 

For this purpose we introduce a simple patch-based convolutional architecture\footnote{Existing patch-based architectures such as MLP designs~\cite{tolstikhin2021MLPMixer,Touvron2021ResMLPFN,ding2021repmlp} or  convMixer~\cite{anonymous2022patches} yield poor accuracy/complexity trade-offs. } that keeps the input resolution constant throughout the network.
This design departs from the historical pyramidal architectures of LeNet~\cite{lecun1989backpropagation}, AlexNet~\cite{Krizhevsky2012AlexNet} or ResNet~\cite{He2016ResNet,He2016IdentityMappings}, to name only a few. 
Their pyramidal design was motivated by the importance of reducing the resolution while increasing the working dimensionality.
That allowed one to maintain a moderate complexity while progressively increasing the working dimensionality, making the space large enough to be separable by a linear classifier.
In our case, we simplify the trunk after a small pre-processing stage that produces the patches. We adopt the same dimensionality throughout all the trunk, fixing it equal to that of the final layer, e.g. our aggregation layer.
We refer to it as \ournet, see Figure~\ref{fig:full_model} for an overview of this network. 

In summary, we make the following contributions: 
\begin{itemize}
    \item We revisit the final pooling layer in convnets by presenting a learned, attention-based pooling;
    
    \item We propose a slight adaptation of our attention-based pooling in order to have one attention map per class, offering a better interpretability of the predictions;
    
    \item We propose an architecture, \ournet{}, with a simple patch-based design (two parameters: depth and width) and a simple training recipe: same learning rate for all our
models, a single regularization parameter. 
\end{itemize}

\noindent We share the architecture definition and pretrained models\footnote{\url{https://github.com/facebookresearch/deit}}. 

%% file: related.tex
\section{Related work}

\paragraph{Attention-based architectures for vision.} 
Early works have introduced attention into convnets~\cite{Bello2019AttentionAC,Ramachandran2019StandAloneSI,Shen2020GlobalSN,Wang2018NonlocalNN,Wu2020VisualTT},  
but it is only recently that a fully attention-based architecture, the vision transformer~\cite{dosovitskiy2020image} (ViT), has become competitive with convnets on ImageNet~\cite{dosovitskiy2020image,Touvron2020TrainingDI}. 
The particularity of this model is that it processes images as a set of non-overlapping patches, without any convolutional or downsampling layers. 
Nevertheless, several works have recently proposed to re-introduce convolutions and downsampling into this architecture. 
For example, some architectures~\cite{graham2021levit,Xiao2021EarlyCH} leverage convolutional layers in the first layers of the vision transformer architecture, while others, such as Swin~\cite{liu2021swin}, LeViT~\cite{graham2021levit}, or PiT~\cite{Heo2021RethinkingSD} exploit a pyramid structure to gradually reduce the spatial resolution of the features. 
These pyramid-based methods are more compatible with prior detection frameworks, and aim at improving the computational efficiency (FLOPs). 
As a downside, these pyramidal approaches dramatically reduce the resolution of the last layers, and hence the quality of their attention maps, making their predictions harder to interpret. Another shortcoming is their relatively high memory usage~\cite{sandler2019nondiscriminative}.

\paragraph{MLP and other patch-based approaches.} 
Architectures based on patches~\cite{liu2021ready}  have been proposed beyond transformers, in particular, based on Multi-Layer Perceptron (MLP) layers such as MLP-Mixer~\cite{tolstikhin2021MLPMixer} and ResMLP~\cite{Touvron2021ResMLPFN}.
Most related to our work, the ablation study of ResMLP~\cite{Touvron2021ResMLPFN} shows the potential of patch-wise convolution over MLPs in terms of performance. 
In line of the ConViT model~\cite{dAscoli2021ConViTIV}, CoatNet~\cite{Dai2021CoAtNetMC} is a patch-based architecture with convolutional blocks followed by transformers blocks.Concurrently, replacing self-attention layers with convolution layers has been explored in ConvMixer~\cite{anonymous2022patches}.

\paragraph{Explainability of the classification decision.} 
There are many strategies to explain the classification decision of a network~\cite{ribeiro2016lime, zeiler2014visualizing}, and most notably by highlighting the most influential regions that led to a decision~\cite{simonyan2014deep, Zhou2016LearningDF,fong2017perturbation}.  
The family of CAM methods~\cite{Wang2020ScoreCAMSV,Chattopadhyay2018GradCAMGG,Selvaraju2019GradCAMVE,Zhou2016LearningDF} shows that the gradients from a network decision contain information about object locations that can be projected back to the image. 
These methods act as general external probes that project the network activity back into the image space, even though Oquab \etal~\cite{oquab2015object} have shown evidence that convnet features contain rough information about the localization of objects.
Unlike these external approaches, the self-attention layers of vision transformers offer a direct access to the location of the information used to make classification decisions~\cite{dosovitskiy2020image, Touvron2020TrainingDI,touvron2021going,caron2021emerging}. 
Our built-in class attention mechanism shares the same spirit of \emph{interpretable by design} computer vision models~\cite{rudin2019nature}. 
However, unlike our mechanism, self-attention layers do not distinguish between classes on the same image without additional steps~\cite{Chefer2021TransformerIB}.

%% file: method.tex
\section{Attention-based pooling with \ournet}
\label{sec:architecture}

\begin{figure*}[ht]
    \centering
     \includegraphics[width=1.0\linewidth,clip,trim=0 0 0 0pt]{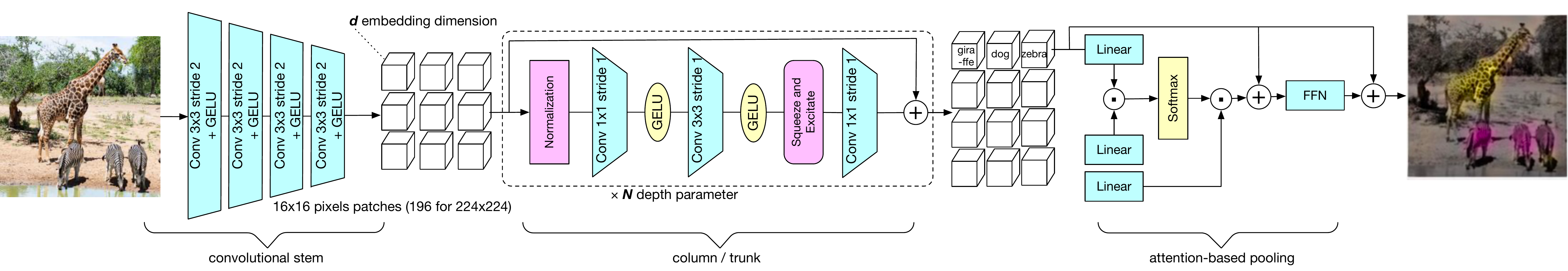}
    \caption{Detail of the full model, with the convolutional stem on the left, the convolutional main block in the middle, and here toppled with multi-class attention-based pooling on the right.
    \label{fig:full_model}}
\end{figure*}

The learned aggregation layer is best associated with a high-resolution feature map. Therefore, while it can be combined with any convolutional architecture like a regular ResNet-50, our suggestion is to combine it with an architecture that maintains the resolution all across the layers. Some works exist, however they offer an underwhelming trade-offs~\cite{tolstikhin2021MLPMixer,Touvron2021ResMLPFN}. To remedy to that problem, we introduce \ournet. This design, which illustrated in Figure~\ref{fig:full_model}, 
is intended to concentrate most of the compute and parameters in the columnar trunk. 
The architecture family is parametrized by the embedding dimension $d$, and the number of repeated blocks in the trunk $N$. 
Below, we describe the architecture and its training in more details.  

\subsection{Architecture design}

\paragraph{The convolutional stem} is a light-weight pre-processing of the image pixels whose role is to segment and map an image into a set of vectors. 
In ViT, this exactly corresponds to the patch extraction step~\cite{dosovitskiy2020image}.
Therefore, we refer to the vectors resulting from this pre-processing as \emph{patches}. 
Recent papers~\cite{graham2021levit,el2021xcit} have shown that it is best to adopt a convolutional pre-processing, in particular for stability reasons~\cite{xiao2021early}. 
In our case, we borrow the convolutional stem from LeVit~\cite{graham2021levit}: a small ConvNet that is applied to the image of size $W\times H \times 3$ and produces a vector map of $W/16 \times H/16 \times d$. 
It can be viewed as a set of $k$ non-overlapping $d$-dimensional patches. 
In our experimental results, except if mentioned otherwise, we use a convolutional stem consisting of four $3\times3$ convolutions with a stride of $2\times2$, followed by a GELU non-linearity~\cite{Hendrycks2016GaussianEL}.
We illustrate the convolutional stem in Figure~\ref{fig:full_model}.

\paragraph{The column,} or trunk, is the part of the model which accounts for most of the layers, parameters, and compute. 
It consists of $N$ stacked residual convolutional blocks as depicted in Figure~\ref{fig:full_model}. 
The block starts with a normalization, followed by a $1\times1$ convolution, then a $3 \times 3$ convolution for spatial processing, a squeeze-and-excitation layer \cite{Hu2017SENet} for mixing channel-wise features, and finally a $1\times1$ convolution right before the residual connection. 
Note that we can interpret the $1\times1$ convolutions as linear layers.
A GELU non-linearity follows the first two convolutions. 
The output of this block has the same shape as its input: the same number of tokens of the same dimension $d$.

Using BatchNorm~\cite{ioffe15batchnorm} often yields better results than LayerNorm~\cite{ba2016layer}, provided the batches are large enough. 
As shown in Section~\ref{sec:experiments}, we also observe this for our model family. 
However, BatchNorm is less practical when training large models or when using large image resolutions because of its dependency on batch size.
In that setup, using BatchNorm requires an additional synchronization step across multiple machines. 
This synchronization increases the amount of node-to-node communication required per step, and in turn, training time. 
In other situations, like for detection and segmentation, the images are large, limiting the batch size and possibly impacting performance. 
Because of all those reasons, unless stated otherwise, we adopt LayerNorm.

\paragraph{Attention-based pooling.} 
At the output of the trunk, the pre-processed vectors are aggregated using a cross-attention layer inspired by transformers.
We illustrate this aggregation mechanism in Figure~\ref{fig:full_model}. 
A \emph{query} class token attends to the projected patches and aggregates them as a weighted summation. 
The weights depend on the similarity of projected patches with a trainable vector (CLS) akin to a class token. 
The resulting $d$-dimensional vector is subsequently added to the CLS vector and processed by a feed-forward network (FFN). 
As opposed to the class-attention decoder by Touvron \etal\cite{touvron2021going} we use a single block and a single head. 
This drastic simplification has the benefit of avoiding the dilution of attention across multiple channels. 
Consequently, the communication between the class token and the pre-processed patches occurs in a single softmax, directly reflecting how the pooling operator weights each patch. 

We can easily specialize the attention maps per class by replacing the CLS vector with a  $k \times d$ matrix, where each of the $k$ columns is associated with one of the classes. 
This specialization allows us to visualize an attention map for each class, as shown in Figure~\ref{fig:visu_attention}. 
The impact of the additional parameters and resulting FLOPS is minimal for larger models in the family. 
However, this design increases peak memory usage and makes the optimization of the network more complicated.
We typically do that in a fine-tuning stage with a lower learning rate and smaller batch size to circumvent these issues. 
By default, we use the more convenient single class token.

\subsection{Discussion: analysis \& properties}
\label{sec:analysis}

\begin{figure*}[h!]
\vspace{-1.5ex}
\begin{minipage}{0.32\linewidth}
\includegraphics[width=\linewidth]{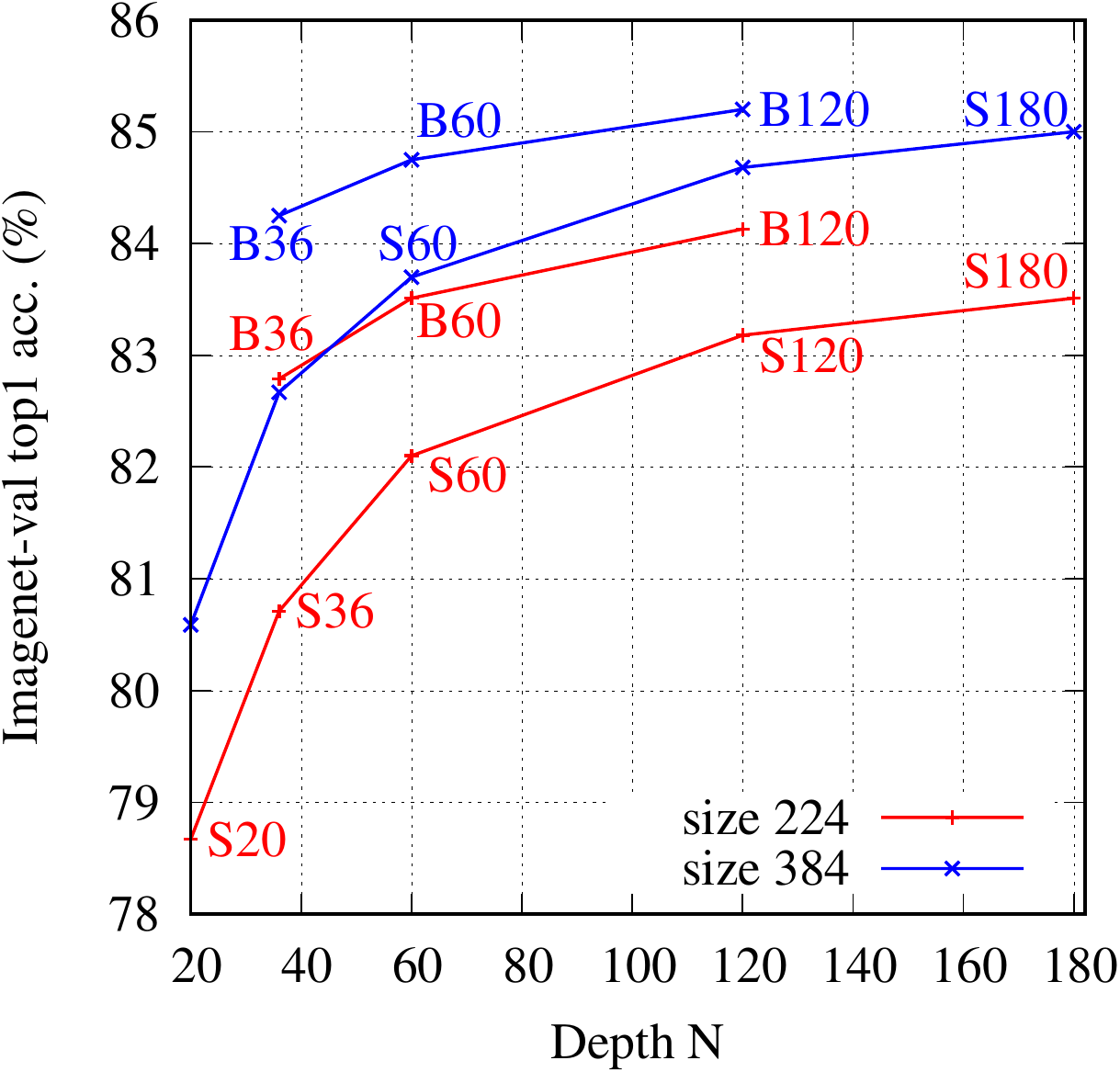}
\caption{Analysis of the accuracy as a function of width (S: $d$\,$=$\,$384$, B: $d$\,=\,$768$) and depth $N$. Depending on the performance criterion (importance of latency, resolution, FLOPs), one could prefer either deeper models or wider models. See Bello \etal ~\cite{Bello2021RevisitingRI} for a study on the relationship between model size, resolution and compute. 
\label{fig:acc_vs_depth_width}} 
\end{minipage}
\hfill
\begin{minipage}{0.32\linewidth}
\includegraphics[width=1.02\linewidth]{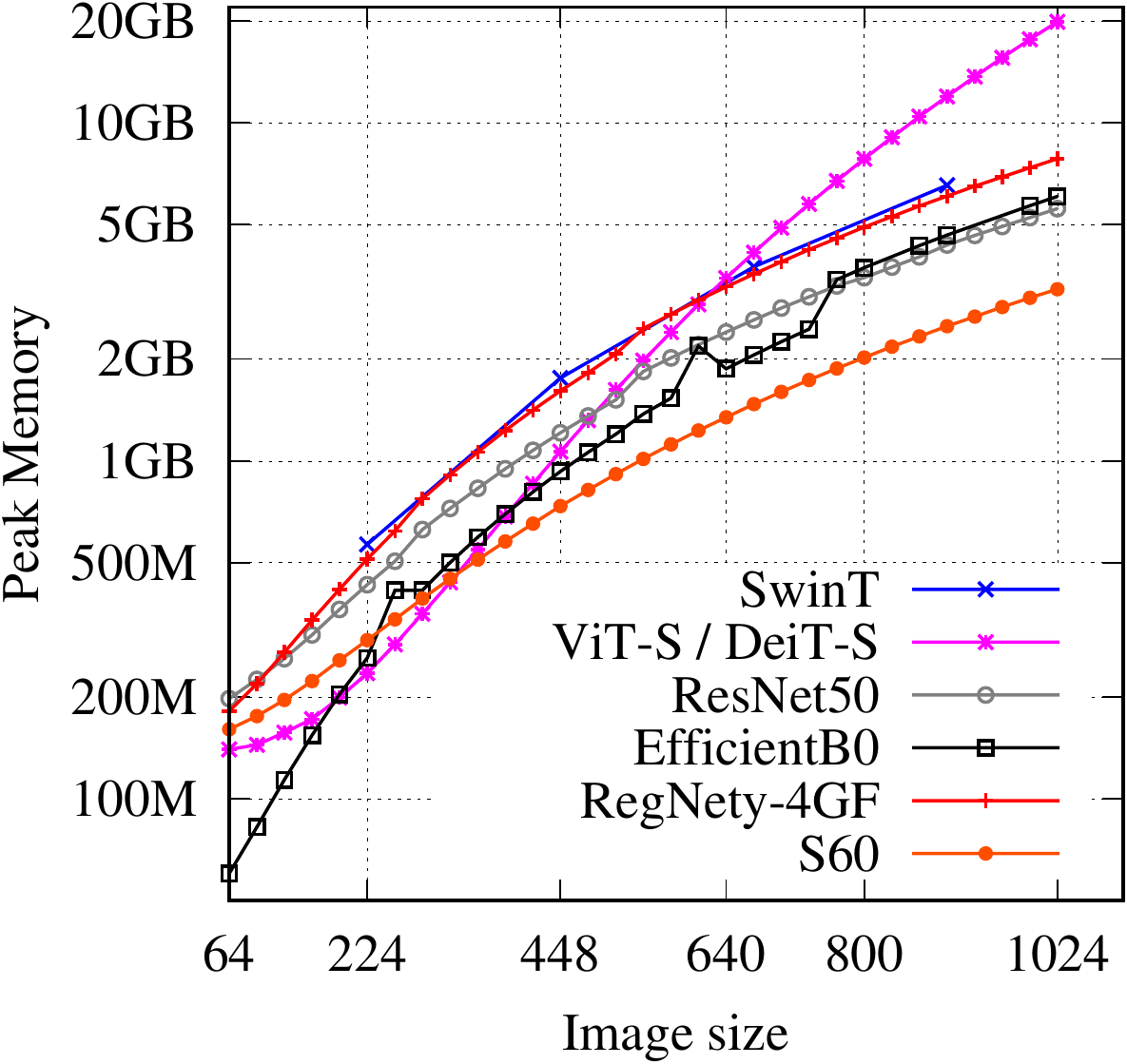}
\caption{Peak memory for varying resolution and different models. Some models like Swin require a full training at the target resolution. Our model scales linearly as a function of the image surface, like other ConvNets. This is in contrast to most attention-based models, which abide by a quadratic complexity for images large enough. 
\label{fig:mem_vs_resolution_S60}} 
\end{minipage}
\hfill 
\begin{minipage}{0.32\linewidth}
\includegraphics[width=\linewidth]{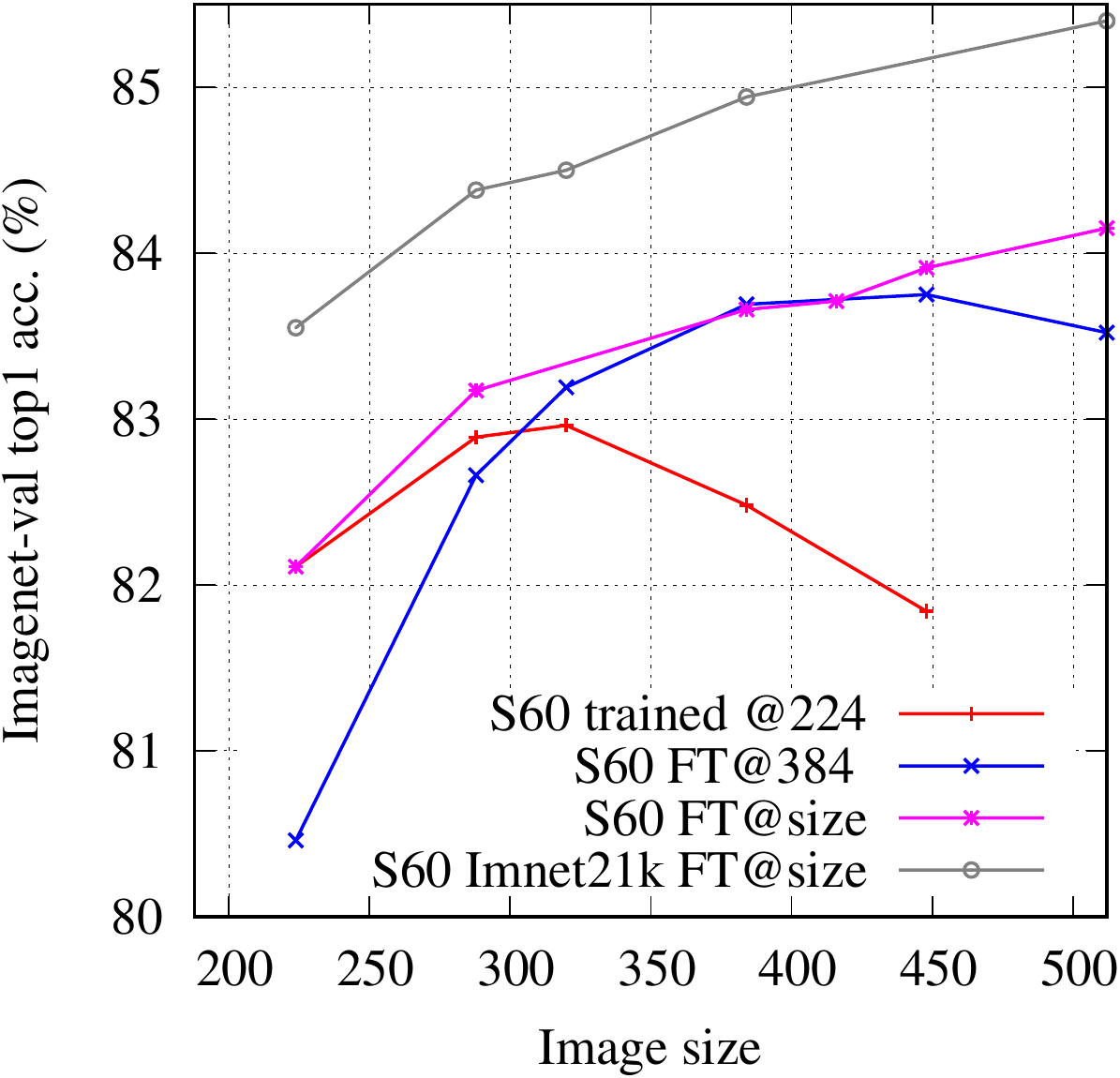}
\caption{Accuracy at different resolutions for the S60 model. We analyze models \textcolor{red}{trained at size 224} or \textcolor{blue}{fine-tuned (FT) @384}, and compare them to \textcolor{magenta}{models fine-tuned at the target inference size} to show the tolerance to test-time resolution changes. 
The \textcolor{black!70}{best model are pre-trained on ImageNet21k} at 224 or 320 and fine-tuned at test-time resolution. 
\label{fig:acc_vs_resolution_S60}} 
\end{minipage}
\vspace{-1.5ex}
\end{figure*}

Below we discuss several properties of our convolutional trunk augmented with the proposed attention-based aggregation stage.

\begin{enumerate}
\item \emph{Simple parametrization.} Our main models are fully defined by width and depth. %
See Figure~\ref{fig:acc_vs_depth_width} for results obtained with these models at two different resolutions (224 and 384). 
Following the same convention as in previous work on vision transformers and vision MLPs~\cite{dosovitskiy2020image,Touvron2020TrainingDI,Touvron2021ResMLPFN}, we refer by S the models with an vector size of $d$\,$=$\,$384$ per patch, by B when $d$\,$=$\,$768$, and by $L$ for $d$\,$=$\,$1024$. 
We use the S60 model for most of our ablations and comparisons since it has a similar number of parameters and FLOPs as a ResNet-50. 

\item \emph{Visualization.} 
Our model allows to easily visualize the network activity. 
Saliency maps are directly extracted from our network without any post-processing. 

\item \emph{Constant resolution across the trunk.} 
The patch-based processing leads to a single processing resolution in the trunk. 
Therefore the activation size is constant across the whole network. 
The memory usage is (almost) constant at inference time, up to the pre- and post-processing stage, which are comparatively less demanding. 
Compared to traditional ConvNets, the network has a coarser processing in the early stages, but a finer resolution towards the output of the trunk. 

\item \emph{Linear scaling with image size.} 
This is a key difference with Vision Transformers. 
Pyramidal transformers such as LeVit, SwinTransformer or MViT partly solve the problem by breaking the quadratic component by rapidly down-scaling the image. 
However, they don't avoid the memory peaks happening with very large images. 
As a consequence of that constant memory usage and linear scaling, our model smoothly scales to larger resolutions, as shown in Figure~\ref{fig:mem_vs_resolution_S60} where we report the Peak Memory usage as a function of the image size.

\item \emph{Easy change of resolution}. 
We do not require any positional encoding, as the relative patch positions are handled by the convolutions. 
In that respect our approach is more flexible than most approaches that needs to be fine-tuned or trained from scratch for each possible target resolution. 
In Figure~\ref{fig:acc_vs_resolution_S60} we show that the properties of our models are quite stable under relatively significant resolution changes. 

\item \emph{No max-pooling.} 
There is no max-pooling or other non-reversible operator in our architecture. 
Formally the function implemented by the trunk is bijective until the aggregation stage. 
We do not exploit this property in this paper, but it may be useful in contexts like image generation~\cite{Donahue2019LargeSA,Kingma2018GlowGF}. 

\end{enumerate}

\subsection{Training recipes} 
\label{sec:training_recipes}

Like many other works (see Liu et al. \cite{liu2021survey}, Table I), our training algorithm inherits from the DeiT~\cite{touvron2021going} procedure for training transformers. 
We adopt the Lamb optimizer~\cite{you20lamb} (a variant of AdamW~\cite{Loshchilov2017AdamW}) with a half-cosine learning schedule and label smoothing \cite{Szegedy2016RethinkingTI}.
For data augmentation, we include the RandAugment~\cite{Cubuk2019RandAugmentPA} variant by Wightman \etal~\cite{wightman2021resnet}, Mixup~\cite{Zhang2017Mixup} ($\alpha=0.8$) and CutMix~\cite{Yun2019CutMix} ($\alpha=1.0$). 
Notably, we include Stochastic Depth~\cite{Huang2016DeepNW} that is very effective for deep transformers~\cite{touvron2021going}, and for which we observe the same effect with our deep \ournet.  
We adopt a uniform drop rate for all layers, and we cross-validate this parameter on ImageNet1k for each model (scores in  Table~\ref{tab:layernorm_vs_batchnorm}). 
We also adopt LayerScale~\cite{touvron2021going}. 
For the deepest models, the drop-rate hyper-parameter (often called ``drop-path'') can be set as high as 0.5, meaning that we can potentially drop half of the trunk. 
A desirable byproduct of this augmentation is that it accelerates the training. 
Note that we do not use gradient clipping, Polyak averaging, or erasing to keep our procedure simple enough. 

We now detail some context-dependent adjustments, based on datasets (ImageNet1k or ImageNet21k), and training (from scratch or fine-tuned). Note that, apart our sensivity study, we use the same Seed 0 for all our experiments~\cite{wightman2021resnet} to prevent picking a ``lucky seed''~\cite{picard21luckyseed} that would not be representative of the model performance.

\paragraph{Training on ImageNet1k.} We train during 400 epochs with a batch size of 2048 and a learning rate fixed at $3.10^{-3}$ for \emph{all models}.  Based on early experiments, we fixed the weight decay to 0.01 for S models and 0.05 for wider models, but practically we observed that the stochastic depth parameter had a preponderant  influence and the most important to adjust, similar to prior observations with ViT \etal~\cite{touvron2021going}. 
We use repeated augmentation~\cite{berman2019multigrain} only when training with this dataset.  %

\paragraph{Fine-tuning at higher resolutions.} We fine-tune our models at higher resolutions in order to correct the train-test resolution discrepancy~\cite{Touvron2019FixRes}, and to analyze the behavior of our models at higher resolutions. This can save a significant amount of resources because models operating at larger resolutions are very demanding to train. 
For fine-tuning, we use a smaller batch size of 1024 in order to compensate for the larger memory requirements. We fix the learning rate to $10^{-5}$, the weight decay to 0.01, and fine-tune during 10 epochs for all our models. 

\paragraph{Training on ImageNet21k.} We train during 90 epochs as in prior works~\cite{dosovitskiy2020image,liu2021swin}. We trained with a batch size of 2048 with a learning rate of $3.10^{-3}$ and weight decay of 0.01, or when possible with a batch size of 4096 to accelerate the training. In that case we adjust the learning rate to $4.10^{-3}$. 

\paragraph{Fine-tuning from ImageNet21k to ImageNet1k} is a more involved modification of the network than just fine-tuning across resolutions because one needs to re-learn the classifiers. In that case, we adopt a longer fine-tuning schedule of 100 epochs %
along with a batch size of 1024 and an initial learning rate of $5.10^{-4}$ with a half-cosine schedule.

%% file: experiments.tex
\section{Main experimental results}
\label{sec:experiments}

This section presents our main experimental results in Image classification, detection and segmentation. We also include an ablation study. 
We refer the reader to the supplemental material for some additional hyper-parameter studies. 
Our code depend on the PyTorch~\cite{pytorch} and timm libraries~\cite{pytorchmodels}. We will share model weights along with a PyTorch implementation of our main models. 

\subsection{Classification results}
We first compare our model with competing approaches on the validation set of ImageNet1k (Imnet-val / Top-1) and ImageNet-v2 in Table~\ref{tab:mainres}. 
We report the compute requirement as reflected by FLOPs, the Peak memory usage, the number of parameters, and a throughput at inference time measured for a constant batch-size of 256 images.

We compare with various models, including classic models like ResNet-50 revisited with modern training recipes such as the one recently proposed by Wightman~\etal~\cite{wightman2021resnet}. 
Note however that different models may have received a different optimization effort, therefore the results on a single criterion are mostly indicative. That being pointed out, we believe that the \ournet results show that a simple columnar architecture is a viable choice compared to other attention-based approaches that are more difficult to optimize or scale.

\begin{figure}[t]
     \includegraphics[height=0.5\linewidth,clip,trim=0 0 0 0pt]{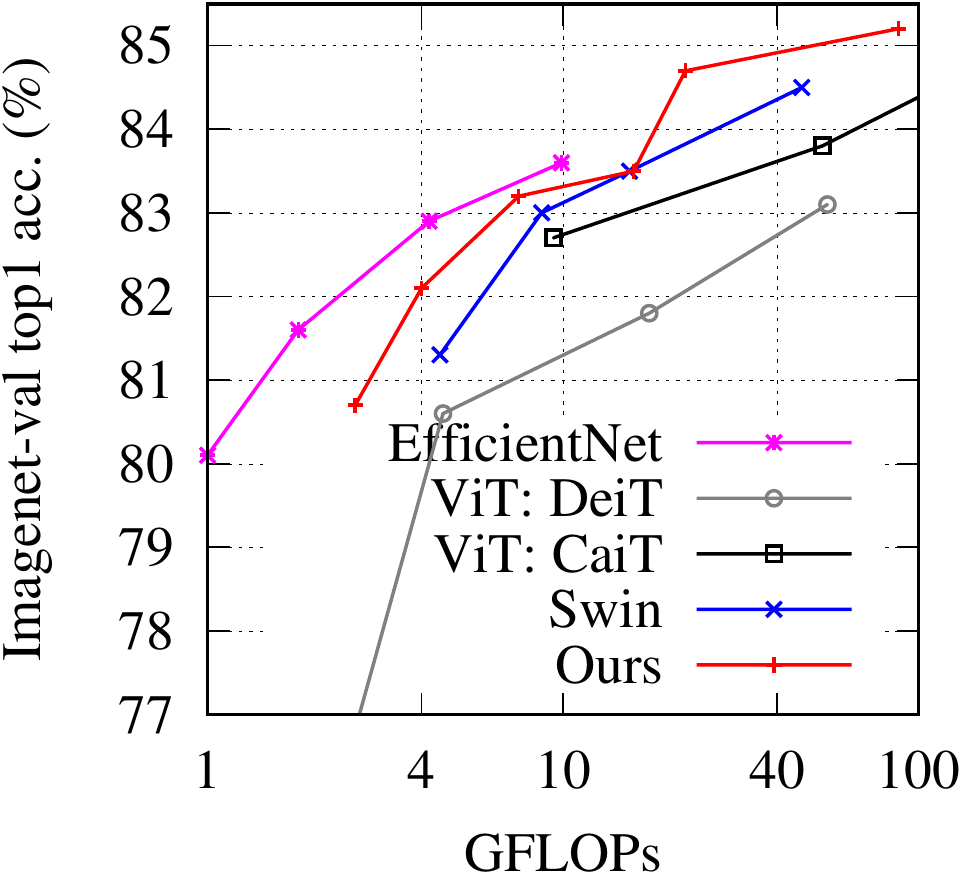}\hfill
     \includegraphics[height=0.5\linewidth,clip,trim=35 0 0 0pt]{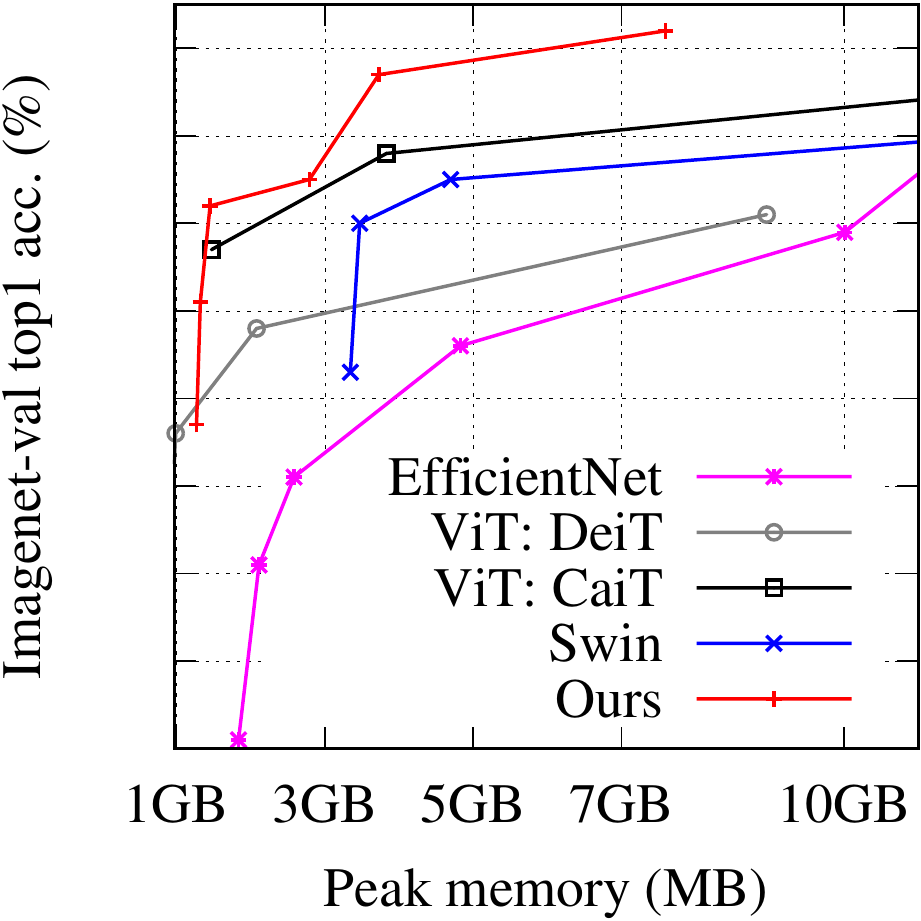}
    \caption{\textbf{Trade-offs} for ImageNet-1k top 1 accuracy vs. FLOPs requirement and peak memory requirements (for a batch of 256 images). %
    Patch-based architectures are comparatively inferior w.r.t. the accuracy-FLOP trade-off than hierarchical ones, but offer better operating points in terms of the accuracy-memory compromise at inference time. 
    \label{fig:acc_flops_memory}}
\end{figure}

\begin{table}[t]

    \caption{
\textbf{Classification with Imagenet1k training.} 
We compare architectures with  based on convolutional networks, Transformers and feedforward networks with comparable FLOPs and number of parameters. All models are trained on ImageNet1k only without distillation nor self-supervised pre-training.
We report Top-1 accuracy on the validation set of ImageNet1k and ImageNet-V2 with different measure of complexity: throughput, FLOPs, number of parameters and peak memory usage. 
The throughput and peak memory are measured on a single V100-32GB GPU with batch size fixed to 256 and mixed precision. 
For ResNet~\cite{He2016ResNet} and RegNet~\cite{Radosavovic2020RegNet} we report the improved results from Wightman et al.~\cite{wightman2021resnet}. Note that different models may have received a different optimization effort. $\uparrow$R indicates that the model is fine-tuned at the resolution $R$. %
\label{tab:mainres}}
\vspace{-1ex}
    \centering
    \scalebox{0.66}{
    \begin{tabular}{@{\ }l@{}c@{\ \ }c@{\ \ \ }r@{\ \ }r|cc@{\ }}
        \toprule
        Architecture        & nb params & throughput & FLOPs & Peak Mem & Top-1  & V2 \\
                      & ($\times 10^6$) & (im/s) & ($\times 10^9$) & (MB)\ \ \ \  & Acc.  & Acc. \\[3pt]
\toprule
\multicolumn{7}{c}{\textbf{``Traditional'' ConvNets}} \\[3pt]
     ResNet-50~\cite{He2016ResNet,wightman2021resnet} &  25.6    & 2587  &  4.1      & 2182 & 80.4  & 68.7 \\
    \midrule
	 RegNetY-4GF~\cite{Radosavovic2020RegNet,wightman2021resnet}       & 20.6  & 1779  & \tzo4.0 & 3041 & 81.5  & 70.7 \\
	 RegNetY-8GF~\cite{Radosavovic2020RegNet,wightman2021resnet}       & 39.2  & 1158 & \tzo8.0 & 3939 & 82.2 & 71.1 \\
	 RegNetY-16GF~\cite{Radosavovic2020RegNet,Touvron2020TrainingDI}      & 83.6  & \pzo714 & \dzo16.0 & 5204   & 82.9  & 72.4 \\

    \midrule
	 EfficientNet-B4~\cite{tan2019efficientnet} & 19.0  & \pzo573 & \tzo4.2 &  10006  & 82.9  & 72.3\\
	 EfficientNet-B5~\cite{tan2019efficientnet} & 30.0  & \pzo268 & \tzo9.9 &  11046  & 83.6  & 73.6\\
	 \midrule
	 NFNet-F0~\cite{Brock2021HighPerformanceLI} & 71.5 & \pzo950 & 12.4 & 4338 & 83.6  & 72.6 \\
	 NFNet-F1~\cite{Brock2021HighPerformanceLI} & 132.6\pzo & \pzo337 & 35.5 & 6628 & 84.7  & 74.4\\
\toprule
\multicolumn{7}{c}{\textbf{Vision Transformers and derivatives}} \\ [5pt]

    ViT: DeiT-S~\cite{Touvron2020TrainingDI,wightman2021resnet}  & 22.0  & 1891 & \tzo4.6 & 987 & 80.6 &  69.4\\
	ViT: DeiT-B~\cite{Touvron2020TrainingDI}    & 86.6  & \pzo831  & \dzo17.5 & 2078 & 81.8 &  71.5\\
	\midrule
	Swin-T-224~\cite{liu2021swin} & 28.3 & 1109 & 4.5 & 3345 & 81.3 &  69.5 \\
    Swin-S-224~\cite{liu2021swin} & 49.6 & \pzo718 & 8.7 & 3470 &  83.0 &   71.8 \\

    Swin-B-224~\cite{liu2021swin} & 87.8  & \pzo532 & 15.4 & 4695 & 83.5 &   \_ \\
    \toprule
\multicolumn{7}{c}{\textbf{Vision MLP}} \\[3pt]

    Mixer-L/16~\cite{tolstikhin2021MLPMixer} &  208.2\pzo &  322 & 44.6 &   2614 &   71.8 & 56.2  \\
    Mixer-B/16~\cite{tolstikhin2021MLPMixer} &  59.9 &   993 & 12.6  & 1448  &   76.4 &  63.2 \\
    ResMLP-S24~\cite{Touvron2021ResMLPFN} &  30.0  &  1681    & \tzo6.0  & 844 &  79.4 &  67.9 \\
    ResMLP-B24~\cite{Touvron2021ResMLPFN} &  116.0\pzo &      1120    & \dzo23.0 & 930  &   81.0 &  69.0 \\
    \toprule
    \multicolumn{7}{c}{\textbf{Patch-based ConvNets}} \\[3pt]
    ResMLP-S12 conv3x3~\cite{Touvron2021ResMLPFN} &  16.7  & 3217 & \tzo3.2  & 763 &  77.0  & 65.5\\
    ConvMixer-768/32~\cite{anonymous2022patches} & 21.1 & \pzo271 & 20.9 & 2644 & 80.2  & \_\\
    ConvMixer-1536/20~\cite{anonymous2022patches} & 51.6 & \pzo157 & 51.4 & 5312 & 81.4  & \_\\
    \midrule
    \rowcolor{Goldenrod}
    \ours-S60 & 25.2  & 1125 & 4.0 & 1321 & 82.1 &  71.0\\
    \rowcolor{Goldenrod}
    \ours-S120&  47.7 & \pzo580 & 7.5 & 1450 & 83.2 &  72.5\\
    \rowcolor{Goldenrod}
    \ours-B60 &  99.4 & \pzo541 & 15.8 & 2790 & 83.5 &   72.6\\
    \rowcolor{Goldenrod}
    \ours-B120 & 188.6\pzo  & \pzo280 & 29.9  & 3314 & 84.1 & 73.9\\
    \bottomrule
    \end{tabular}}
\end{table}

\paragraph{Higher-resolution.} 
There is a fine interplay between model size and resolution when it comes to the specific optimization of FLOPs and accuracy. We refer to the findings of Bello \etal~\cite{Bello2021RevisitingRI} who discussed some of these relationships, for instance the fact that small networks are better associated with smaller resolution. In our work we have not  optimized for the Pareto curve specifically. Since this trade-off is only one out of multiple criteria depending on the context, we prefer to report most of our results at the 224 and 384 resolutions. Table~\ref{tab:mainres} shows that our model significantly benefit from larger resolution images. See also Figures~\ref{fig:mem_vs_resolution_S60}  and \ref{fig:acc_vs_resolution_S60} where we analyze \ournet as a function of the image size. 
Table~\ref{tab:higher_res} we analyze \ournet pre-trained on ImageNet21k with different fine-tuning resolution. All network are pre-trained on ImageNet21k during 90 epochs at resolution $224\times 224$, finetune on ImageNet1k at resolution $384\times 384$ and then fine-tune at bigger resolution. 
 
\begin{table}[t]
    \centering
    \caption{\textbf{ImageNet21k pre-training:} Comparison of \ournet fine-tuned at different resolutions on ImageNet1k. We report peak memory (MB) and throughput (im/s) on one GPU V100 with batch size 256 and mixed precision. Larger resolution provides classification improvement with the same model, but significantly increase the resource requirements.  
    [\emph{italic refers to a few results obtained with a longer training}]. 
    }
    \vspace{-1ex}
    \scalebox{0.7}{
    \begin{tabular}{cccccc}
    \toprule
         Model & GFLOPs & Peak Mem & throughput & Res & Imnet-val Acc  \\
         \midrule
         S60   &   \pzo\pzo4.0      &   \pzo1322        &    1129      &  224  & \phantom{[{\emph{83.3}}]} 82.9 [{\emph{83.5}}]  \\
         S60   &     \pzo\pzo6.6    &    \pzo2091       &    \pzo692      &  288  & \phantom{[{\emph{84.2}}]} 84.0 [{\emph{84.4}}]  \\
         S60   & \pzo11.8    & \pzo3604  & \pzo388  &  384 & \phantom{[\emph{85.0}]} 84.6 [{\emph{84.9}}] \\
         S60   & \pzo20.9    & \pzo6296  & \pzo216 &  512 & \phantom{[{\emph{85.4}}]} 85.0 [{\emph{85.4}}]\\
         \midrule
         B60   &   \pzo15.8      &  \pzo2794        &    \pzo547      &  224  &  \phantom{[{\emph{85.0}}]} 85.0 [{\emph{85.4}}]  \\
         B60   &   \pzo26.1      &   \pzo4235        &    \pzo328      &  288  & 85.7  \\
         B60   & \pzo46.5    & \pzo7067  & \pzo185 &  384 &  \phantom{[{\emph{86.5}}]} 86.1   [{\emph{86.5}}]\\

         \midrule 
         L60   &   \pzo28.1      &   \pzo3913        &     \pzo394     &  224  &  85.6  \\
         L60   &   \pzo46.4      &    \pzo5801       &    \pzo237      &  288  &  86.1  \\
         L60   &   \pzo82.5      &    \pzo9506       &   \pzo132       &  384  &  86.4  \\
         \midrule
         B120   &  \pzo29.8       &   \pzo3313        &     \pzo280     & 224  & 86.0    \\
         B120   &   \pzo49.3      &    \pzo4752       &     \pzo169     & 288  & 86.6    \\
         B120   &  \pzo87.7   & \pzo7587  & \pzo\pzo96 &  384 & 86.9   \\

         \midrule 
         L120   &   \pzo53.0      &    \pzo4805       &    \pzo204      &  224  &  86.1  \\
         L120   &    \pzo87.5     &    \pzo6693       &    \pzo123      &  288  &  86.6  \\
         L120   &    155.5     &      10409     &    \pzo\pzo68      &  384  &  87.1  \\
        
         \bottomrule
    \end{tabular}
    \label{tab:higher_res}}
\end{table}

\subsection{Segmentation results and detection}

\paragraph{Semantic segmentation}
We evaluate our models with semantic segmentation experiments on the ADE20k dataset~\cite{Zhou2017ScenePT}.
This dataset consist in 20k training and 5k validation images with labels over 150 categories. 
For the training, we adopt the same schedule as in Swin~\cite{liu2021swin}:~160k iterations with UpperNet~\cite{Xiao2018UnifiedPP}. 
At test time we evaluate with a single scale similarly to XciT~\cite{el2021xcit} and multi-scale as in Swin~\cite{liu2021swin}.
As our approach is not pyramidal we only use the final output of our network in UpperNet. 
Unlike concurrent approaches we only use the output of our network at different levels in UpperNet which simplifies the approach.

Our results are reported in Table~\ref{tab:sem_seg}.
We can observe that our approach although simpler is at the same level as the state-of-the-art Swin architecture~\cite{liu2021swin} and outperforms XCiT~\cite{el2021xcit} in terms of FLOPs-mIoU tradeoff.

\begin{table}[t]

       \caption{\textbf{ADE20k semantic segmentation} performance using UperNet \cite{xiao2018unified} (in comparable settings). All models are pre-trained on ImageNet1k except models with $^\dagger$ symbol that are pre-trained on ImageNet21k. 
       \label{tab:sem_seg}}
           \vspace{-1ex}
        \centering
        \scalebox{0.7}{
        \begin{tabular}{lcccc}
        \toprule
             \multirow{2}{*}{Backbone} & \multicolumn{4}{c}{UperNet}  \\
             & \#params & FLOPs & Single scale & Multi-scale  \\
             \cmidrule{2-3}
             \cmidrule{4-5}
             & ($\times 10^6$) & ($\times 10^9$) & mIoU  & mIoU \\
            \midrule 
             ResNet50 \cite{He2016ResNet} & 66.5 & \_ & 42.0 & \_ \\
             DeiT-S \cite{Touvron2020TrainingDI} & 52.0 & 1099 & \_ & 44.0 \\
             XciT-T12/16 \cite{el2021xcit} & 34.2 & \pzo874  & 41.5  & \_\\
             XciT-S12/16 \cite{el2021xcit}  & 54.2  & \pzo966 & 45.9 & \_\\
             Swin-T  \cite{liu2021swin} & 59.9 & \pzo945 & 44.5 & 46.1 \\

             \rowcolor{Goldenrod}
             \ours-S60 & 57.1 & \pzo952 & \textbf{46.0}  & \textbf{46.9}\\

            \midrule
             XciT-M24/16 \cite{el2021xcit} & 112.2 & 1213 & 47.6 & \_ \\
             XciT-M24/8 \cite{el2021xcit} & 110.0 & 2161  & 48.4 & \_ \\
             Swin-B \cite{liu2021swin} & 121.0 & 1188  & 48.1 & 49.7 \\

            \rowcolor{Goldenrod}
            \ours-B60 & 140.6 & 1258  & 48.1  & 48.6 \\

              \rowcolor{Goldenrod}
             \ours-B120 & 229.8 & 1550  & \textbf{49.4}  & \textbf{50.3}\\
            \midrule 
            Swin-B$^\dagger$ ($640\times 640$) & 121.0 & 1841  & 50.0 & 51.6 \\
            CSWin-B$^\dagger$~\cite{Dong2021CSWinTA} & 109.2 & 1941  & 51.8 & 52.6 \\

             \rowcolor{Goldenrod}
             \ours-S60$^\dagger$ & \pzo57.1 & \pzo952 & 48.4  & 49.3 \\
            \rowcolor{Goldenrod}
            \ours-B60$^\dagger$ & 140.6 & 1258  & 50.5  & 51.1 \\
            \rowcolor{Goldenrod}
             \ours-B120$^\dagger$ & 229.8 & 1550  & 51.9  & 52.8  \\

            \rowcolor{Goldenrod}
             \ours-L120$^\dagger$ & 383.7 & 2086  &  \textbf{52.2} & \textbf{52.9}   \\

        \bottomrule     
        \end{tabular}
        } 
\end{table}

\paragraph{Detection \& instance segmentation}

We have evaluated our models on detection and instance segmentation tasks on COCO~\cite{Lin2014MicrosoftCC}. 
We adopt the Mask R-CNN~\cite{he2017mask} setup with the commonly used $\times 3$ schedule.
Similar to segmentation experiments, as our approach is not pyramidal, we only use the final output of our network in Mask R-CNN~\cite{he2017mask}. 
Our results are in Table~\ref{tab:coco_det}.
We can observe that our simple approach is on par with state of the art architecture like  Swin~\cite{liu2021swin} and XCiT~\cite{el2021xcit} in terms of FLOPs-AP tradeoff.

\begin{table}[t]
        \captionof{table}{\footnotesize \textbf{COCO object detection and instance segmentation} performance on the mini-val set. All backbones are pre-trained on ImageNet1k, use Mask R-CNN model~\cite{he2017mask} and are trained with the same 3$\times$ schedule. 
        }%
            \vspace{-1ex}
        \centering
        \scalebox{0.72}{
        \begin{tabular}{l@{\ \ }r c@{\ \ } | c@{\ \ \ }c@{\ \ }c|c@{\ \ }c@{\ \ } c}
        \toprule
             Backbone & \!\!\!\ \#params\!\!\! & \!\!\!GFLOPs\ \!\!\! & $\text{AP}^{b}$ & $\text{AP}^{b}_{50}$ & $\text{AP}^{b}_{75}$ & $\text{AP}^{m}$ & $\text{AP}^{m}_{50}$ & $\text{AP}^{m}_{75}$\\ 
            \midrule 
            ResNet50 \cite{He2016ResNet} & 44.2M & 180 & 41.0 & 61.7 & 44.9 & 37.1 & 58.4 & 40.1 \\
            ResNet101 \cite{He2016ResNet} & 63.2M & 260 & 42.8 & 63.2 & 47.1 & 38.5 & 60.1 & 41.3 \\
            ResNeXt101-64 \cite{xie2017aggregated} & \!101.9M\! & 424 & 44.4 & 64.9 & 48.8 & 39.7 & 61.9 & 42.6 \\
            \midrule
            
            PVT-Small \cite{wang2021pyramid} & 44.1M & \_  & 43.0 & 65.3 & 46.9 & 39.9 & 62.5 & 42.8 \\
            PVT-Medium \cite{wang2021pyramid} & 63.9M & \_ & 44.2 & 66.0 & 48.2 & 40.5 & 63.1 & 43.5 \\

            XCiT-S12/16 & 44.4M & 295 & 45.3 & 67.0 & 49.5 & 40.8 & 64.0 & 43.8  \\
            XCiT-S24/16 \cite{el2021xcit} & 65.8M & 385  & 46.5 & 68.0 & 50.9 & 41.8 & 65.2  & 45.0 \\

            ViL-Small \cite{zhang2021multi} & 45.0M & 218 & 43.4 & 64.9 & 47.0 & 39.6 & 62.1 & 42.4 \\
            ViL-Medium \cite{zhang2021multi} & 60.1M & 294 & 44.6 & 66.3 & 48.5 & 40.7 & 63.8 & 43.7 \\
            ViL-Base \cite{zhang2021multi}  & 76.1M & 365 & 45.7 & 67.2 & 49.9 & 41.3 & 64.4 & 44.5 \\

            Swin-T \cite{liu2021swin} & 47.8M & 267 & 46.0 & 68.1  & 50.3 & 41.6 & 65.1 & 44.9 \\

           \rowcolor{Goldenrod}
            \ours-S60 & 44.9M  & 264 &  46.4 & 67.8 & 50.8 & 41.3 & 64.8 &44.2 \\
            \rowcolor{Goldenrod}
            \ours-S120 & 67.4M &  339 & 47.0 & 69.0  & 51.4 & 41.9 & 65.6 & 44.7\\
        \bottomrule     
     \end{tabular}     
     } %
     \label{tab:coco_det}
\end{table}

\subsection{Ablations}

All our ablation have been carried out with ``Seed 0'', i.e., we report only one result without handpicking. For this reason one must keep in mind that there is a bit of noise in the performance measurements: On ImageNet1k-val, we have measured with the seeds 1 to 10 a standard deviation of $\pm 0.11\%$ in top-1 accuracy for a S60 model, which concurs with measurements done on ResNet-50 trained with modern training procedures~\cite{wightman2021resnet}. 

\begin{figure}
\includegraphics[width=\linewidth]{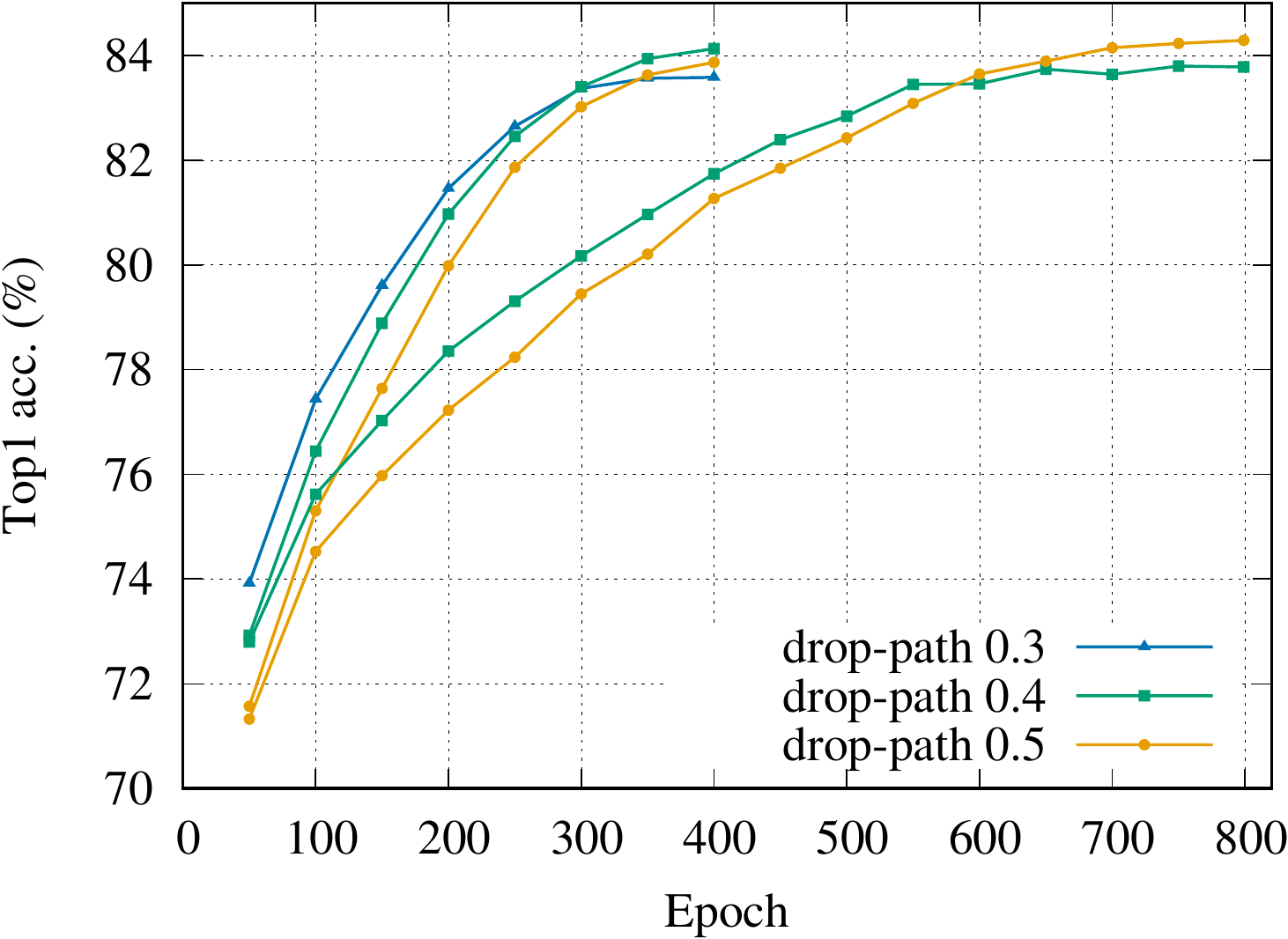}
\caption{Effect of stochastic depth on the performance for  varying training duration for a \ournet-B120 model trained @ resolution 224. 
The corresponding hyper-parameter (\textit{drop-path}) is selected among 0.3, 0.4 or 0.5 in that case, which means that we randomly drop up to half of the layers. Smaller values of the drop-rate converge more rapidly but saturate. 
\label{fig:stochastic_depth}} 
\end{figure}

\paragraph{Stochastic depth.} Our main parameter is the stochastic depth, whose effect is analyzed in Fig. \ref{fig:stochastic_depth}. This regularization slows down the training, yet with long enough schedules, higher values of the \textit{drop-path} hyperparameter lead to better performance at convergence.  
We train with the values reported in Table~\ref{tab:layernorm_vs_batchnorm}. When fine-tuning at higher resolutions or from ImageNet21k, we reduce this \textit{drop-path} by 0.1. 
See also Appendix~\ref{sec:apdx_explo} for a preliminary ablation on the learning rate and weight decay, which showed that the performance is relatively stable with respect to these parameters. Fixing this hyper-parameter couple is possibly suboptimal but makes it convenient and more resource-efficient to adjust a single hyper-parameter per model. Therefore, we have adopted this choice in all our experiments. 

\paragraph{Architectural ablation.} 
In Table~\ref{tab:ablation_archi}, we have conducted various ablations of our architecture with the S60 model.
We compare the impact of class attention \textit{vs.} average-pooling. Average-pooling is the most common aggregation strategy in ConvNet while class attention is only used with transformers~\cite{touvron2021going}. 
We compare also convolutional stem \textit{vs.} linear projection for the patch extraction in the image, 
LayerNorm \textit{vs.} BatchNorm and 
Multi-heads class attention as used in CaiT~\cite{touvron2021going} \textit{vs.} single-head class attention. Our single-head design reduces the memory consumption and simplifies attention map visualization.

\begin{table}[t]
\caption{Ablation of our model: we modify each time a single architectural characteristic in our \ournet model S60, and measure how it affects the classification performance on ImageNet1k. Batch-normalization improves the performance a bit. The convolutional stem is key for best performance, and the class-attention brings a slight improvement in addition to enabling attention-based visualisation properties. 
\label{tab:ablation_archi}}
\vspace{-0.5ex}
\centering
\scalebox{0.8}{
    \begin{tabular}{rclr}
    \toprule
    \multicolumn{3}{l}{$\downarrow$ Modification to the architecture} & Top-1 acc. \\
    \midrule
    \rowcolor{blue!5} 
    none    &  & &   82.1   \\
    class-attention       & $\rightarrow$ & average pooling  & 81.9\\ 
    conv-stem             & $\rightarrow$ & linear projection & 80.0 \\ 
    layer-normalization   & $\rightarrow$ & batch-normalization & 82.4 \\ 
    single-head attention & $\rightarrow$ & multi-head attention & 81.9 \\ 
    a single class-token  & $\rightarrow$ & one class-token per class & 81.1 \\ 
    \bottomrule 
    \end{tabular}}
\end{table}

\paragraph{Attention-based pooling with ConvNets.}

Interestingly, our learned aggregation stage increases the performance of a very competitive ResNet model. When adopting the recent training recipe from Wightman \etal~\cite{wightman2021resnet}, %
we obtain $80.1\%$ top-1 accuracy on Imagenet1k by adding a learned pooling to a ResNet50. This is an improvement of $+0.3\%$ to the corresponding 300-epoch baseline based on average pooling. 
The class attention  only slightly increases the number of FLOPs of the models:
4.6B vs 4.1B. %

We point out that we have not optimized the training recipes further (either without or with class-attention). This result is reported for a single run (Seed 0) in both cases.

\paragraph{Patch pre-processing.} 
In the vanilla patch-based approaches as vision transformers~\cite{dosovitskiy2020image,Touvron2020TrainingDI} and MLP-style models~\cite{tolstikhin2021MLPMixer,Touvron2021ResMLPFN}, the images patches are embedded by one linear layer.
Recent works~\cite{graham2021levit,Xiao2021EarlyCH} show that replacing this linear patch pre-processing by a few convolutional layers allows to have a more stable architecture~\cite{Xiao2021EarlyCH} with better performance.
So, in our work we choose to use a convolutional stem instead of pure linear projection.
We provide in Table~\ref{tab:ablation_archi} an ablation of this component.

%% file: conclusion.tex
\section{Conclusion}

In this paper, we introduced a full patch-based ConvNet with no pyramidal structure. %
We used an attention-based pooling on top of the trunk, akin to the attention mechanism in transformers, which offers visualization properties. 
Our model is only parametrized by its width and depth, and its training does not require a heavy hyper-parameter search. %
We demonstrated its interest on several computer vision tasks: classification, segmentation, detection.
\bigskip

\noindent\textbf{Limitations:\,} There is no perfect metric for measuring the overall performance of a given neural network architecture~\cite{Dehghani2021TheEM}. 
We have provided 4 different metrics but there are probably some aspects that are not considered.
Deep and wide models have the same behaviour with respect to FLOPs but the wider models have the advantage to be associated with a lower latency~\cite{Goyal2021NondeepN,Zagoruyko2016WideRN}.
We have mostly experimented with depth rather than width because deep models consume less memory at inference time, which makes them an appealing choice when dealing with higher resolution images~\cite{Bello2021RevisitingRI}, as is the case in segmentation and detection. 

\bigskip

\noindent\textbf{Broader impact:\,} 
Large scale deep learning models are effective for many different computer vision applications, but the way they reach their decision is still not yet fully understood. When deploying such machine learning-based systems, there would be a benefit to be able to illustrate their choices  in critical applications. We hope that our model, by its simplicity, and by its built-in internal visualization mechanism, may foster this direction of interpretability.

%% file: appendix.tex
\appendix\newpage

\input{supplementary_title}

\section{Hyper-parameter study: exploration phase}
\label{sec:apdx_explo}
In this appendix we discuss the grid searches that we have done for the material presented in this paper. 
During our exploration phase, we have modified only a few variables hyper-parameters to avoid some potential overfitting, which usually results from the exploration of a large hyper-parameter space: we have solely changed the learning rate (LR), the weight decay (WD) and the drop-path parameter involved in stochastic depth (SD). 
For the same reason we have selected a relatively coarse grid search.
We have fixed the batch size to 2048, and changed the hyper-parameters by setting them from the following values:
\begin{itemize}
    \item LR $\in$ \{\ 0.001, 0.0015, 0.002, 0.003, 0.004, 0.005 \} ;
    \item WD $\in$ \{\ 0.001, 0.01, 0.03, 0.05, 0.1, 0.15, 0.2 \} ; 
    \item SD $\in$ \{\ 0, 0.05, 0.1, 0.15, 0.2, 0.3, 0.4 ,0.5 \}.
\end{itemize}

Note, we have not exhaustively spanned the product space of these values with a grid search: after a few tests on a few models (mostly: S36 \& S60), we concluded that we could set LR\,=\,$3.10^{-3}$. We had the same conclusion for setting WD\,=\,0.01, yet for larger models trained on Imagenet-val, we preemptively increased the regularization to WD=0.05 for larger models  ($d$\,=\,$384$) in case the lack of regularization would have affected the convergence (which we  noticed with very small values of WD for small models, see our ablation in Table~\ref{tab:sensitivity_hparams}). However, the difference does not seem statistically significant from the value WD\,=\,0.01 in the few experiments that we have done subsequently. While our choice are likely not optimal for all models, in our opinion the benefit of taking a single tuple (LR,WD) for models of all depth vastly overcome the risk of overfitting/over-estimating the performance. 
The other hyper-parameters are inherited from typical values in the literature~\cite{Touvron2020TrainingDI,wightman2021resnet} without any optimization from us, and therefore could potentially be improved. 

Regarding the last hyper-parameter SD, as observed by Touvron et al.~\cite{touvron2021going} for vision transformers, we noticed that validating this hyper-parameter properly is key to performance. Since this validation is carried out on Imagenet, in the main paper we have reported results on Imagenet-V2 to ensure an independent test set.

\section{Ablations}

\paragraph{Hyper-parameters. }
Table~\ref{tab:sensitivity_hparams} and~\ref{tab:sensitivity_sd} provide the accuracy obtained when varying our hyper-parameters with the S60 model, with our baseline as LR\,=\,$3.10^{-3}$, WD\,=\,0.01 and SD\,=\,0.15.

\begin{table}[h]
\centering
\scalebox{0.8}{
\begin{tabular}{lllll}
\toprule
Model & LR & WD & SD & Imagenet-val \\
\toprule
\multicolumn{5}{c}{\textbf{ablation: learning rate}} \\[5pt]
S60	&  0.0005  & 0.01  &  0.15  &  \quad 77.00  \\
S60	&  0.0010  & 0.01  &  0.15  &  \quad 80.70  \\
S60	&  0.0015  & 0.01  &  0.15  &  \quad 81.58  \\
S60	&  0.0020  & 0.01  &  0.15  &  \quad 81.92  \\
\rowcolor{blue!5}  
S60	&  0.0030  & 0.01  &  0.15  &  \quad 82.10  \\
S60	&  0.0040  & 0.01  &  0.15  &  \quad 81.59  \\
S60	&  0.0050  & 0.01  &  0.15  &  \quad 80.31  \\
\rowcolor{red!5}  
S60	&  0.0070  & 0.01  &  0.15  &  \quad failed@34  \\
\toprule
\multicolumn{5}{c}{\textbf{ablation: weight decay}} \\[5pt]
\rowcolor{red!5}  
S60  & 0.0030  &  0.001  &  0.15  &  \quad failed@92   \\ 
\rowcolor{red!5}  
S60  & 0.0030  &  0.002  &  0.15  &  \quad failed@105  \\  
S60  & 0.0030  &  0.005  &  0.15  &  \quad 81.66    \\
\rowcolor{blue!5} 
S60  & 0.0030  &  0.010  &  0.15  &  \quad 82.10    \\
S60  & 0.0030  &  0.020  &  0.15  &  \quad 82.03    \\
S60  & 0.0030  &  0.050  &  0.15  &  \quad 81.59    \\
S60  & 0.0030  &  0.100  &  0.15  &  \quad 81.33    \\
\bottomrule
\end{tabular}}
\caption{Sensitivity to our hyper-parameters for the S60 model: Learning Rate (LR), Weight Decay (WD). Rows highlight in red with ``fail@E'' indicates that the training has failed at Epoch E. The model reaches a reasonable performance over a wide set of values. For instance the intervals $\mathrm{LR} \in [0.2,0.3]$ or $\mathrm{WD} \in [0.01,0.02] $ lead to similar values. The optimization is stable with  reasonable performance for hyper-parameters covering large intervals ($\mathrm{LR} \in [0.1,0.5]$ or $\mathrm{WD} \in [0.005,0.05]$). 
\label{tab:sensitivity_hparams}} 
\end{table}

Some regularization is needed for convergence and the learning rate should be kept below a threshold (0.005). 
The LR and SD hyper-parameters are the more influential on the performance. Table~\ref{tab:sensitivity_sd} analyses their interaction, which shows that they can be set relatively independently.  

\begin{table}[h]
\centering
\scalebox{0.8}{
\begin{tabular}{l|cccc}
\toprule
 & \multicolumn{4}{c}{\textbf{learning rate}} \\[5pt]
SD			&   0.001  & 0.0015   & 0.002  &   0.003  \\
\midrule	
	0   	&	79.51  & 80.01	  & 80.56  &   80.77  \\
	0.05	&	80.62  & 81.56	  & 81.60  &   81.82  \\
	0.1  	&	80.75  & 81.78	  & 82.00  &   81.90  \\
	0.15	&	80.70  & 81.58	  & 81.92  &   \cellcolor{blue!5}82.10  \\
	0.2 	&	80.43  & 81.44	  & 81.70  &   81.90  \\
\bottomrule
\end{tabular}}
\caption{Analysis of Learning rate vs stochastic depth hyper-parameters  (S60, WD=0.01). 
\label{tab:sensitivity_sd}} 
\end{table}

\paragraph{LayerNorm vs BatchNorm.}
LayerNorm is the most used normalisation in transformers while BatchNorm is the most used normalisation with ConvNets.
For simplicity we have used LayerNorm as it does not require (batch) statistics synchronisation during training, which tends to significantly slow the training, especially on an infrastructure with relatively high synchronisation costs.

In Table~\ref{tab:layernorm_vs_batchnorm}
we compare the effects of LayerNorm with those of  BatchNorm. 
We can see that BatchNorm increases the \ournet top-1 accuracy. This difference tends to be lower for the deeper models.

\begin{table}[t]
\caption{Comparison of \ournet with Layer-Normalization and Batch-Normalization: Performance on Imagenet-1k-val after pre-training on Imagenet-1k-train only. The \textit{drop-path} parameter value is obtained by cross-validation on Imagenet1k for each model. Batch-Normalization usually provides a slight improvement in classification, but 
but with large models the need to synchronization can significantly slow down the training (in  some cases like training a B120 model on AWS, it almost doubled the training time). 
Therefore we do not use it in the main paper. 
\label{tab:layernorm_vs_batchnorm}}
\vspace{-0.5ex}
\centering
\scalebox{0.8}{
    \begin{tabular}{lccc}
    \toprule
    & & \multicolumn{2}{c}{Imagenet-val Top-1 acc.}   \\
    \cmidrule(l){3-4}
    Model  & \textit{drop-path} & LayerNorm  & BatchNorm  \\ 
    \midrule   
S20	& 0.0\pzo  & 78.7	&  78.8  \\
S36	& 0.05     & 80.7	&  81.2  \\
S60	& 0.15     & 82.1	&  82.4  \\
S120& 0.2\pzo  & 83.2	&  83.4  \\
B36 & 0.2\pzo  & 82.8	&  83.5  \\
B60 & 0.3\pzo  & 83.5	&  83.9  \\
B120& 0.4\pzo  & 84.1	&  84.3  \\
    \bottomrule 
    \end{tabular}}
\end{table}

\section{Additional results}

\begin{table}[h!]

    \caption{
\textbf{Comparison of architectures on classification.}
We compare different architectures  based on convolutional networks, Transformers and feedforward networks with comparable FLOPs and number of parameters. All models are trained on ImageNet1k only without distillation nor self-supervised pre-training.
We report Top-1 accuracy on the validation set of ImageNet1k and ImageNet-V2 with different measure of complexity: throughput, FLOPs, number of parameters and peak memory usage. 
The throughput and peak memory are measured on a single V100-32GB GPU with batch size fixed to 256 and mixed precision. 
For ResNet~\cite{He2016ResNet} and RegNet~\cite{Radosavovic2020RegNet} we report the improved results from Wightman et al.~\cite{wightman2021resnet}. Note that different models may have received a different optimization effort. $\uparrow$R indicates that the model is fine-tuned at the resolution $R$.
\label{tab:mainres_ext}}
\vspace{-1ex}
    \centering
    \scalebox{0.66}{
    \begin{tabular}{@{\ }l@{}c@{\ \ }c@{\ \ \ }r@{\ \ }r|cc@{\ }}
        \toprule
        Architecture        & nb params & throughput & FLOPs & Peak Mem & Top-1  & V2 \\
                      & ($\times 10^6$) & (im/s) & ($\times 10^9$) & (MB)\ \ \ \  & Acc.  & Acc. \\[3pt]

\toprule
\multicolumn{7}{c}{\textbf{``Traditional'' ConvNets}} \\[3pt]
     ResNet-50~\cite{He2016ResNet,wightman2021resnet} &  25.6    & 2587  &  4.1      & 2182 & 80.4  & 68.7 \\
    \midrule
	 RegNetY-4GF~\cite{Radosavovic2020RegNet,wightman2021resnet}       & 20.6  & 1779  & \tzo4.0 & 3041 & 81.5  & 70.7 \\
	 RegNetY-8GF~\cite{Radosavovic2020RegNet,wightman2021resnet}       & 39.2  & 1158 & \tzo8.0 & 3939 & 82.2 & 71.1 \\
	 RegNetY-12GF~\cite{Radosavovic2020RegNet,wightman2021resnet}      & 52  & 835.1 & \dzo12.0 & 5059 & \\
	 RegNetY-16GF~\cite{Radosavovic2020RegNet,Touvron2020TrainingDI}      & 83.6  & \pzo714 & \dzo16.0 & 5204   & 82.9  & 72.4 \\
	 RegNetY-32GF~\cite{Radosavovic2020RegNet,wightman2021resnet}      & 145  & 441.7  & \dzo32.0 & 5745.4   & \\

    \midrule
    EfficientNet-B0~\cite{tan2019efficientnet} & 5.3  &3856  & \tzo0.4  &  1835  &77.1  & 64.3 \\
    EfficientNet-B1~\cite{tan2019efficientnet} &7.8   & 2450 & \tzo0.7  &  2111  & 79.1  & 66.9\\
    EfficientNet-B2~\cite{tan2019efficientnet} & 9.2   & 1851 & \tzo1.0 & 2584   & 80.1  & 68.8 \\
	 EfficientNet-B3~\cite{tan2019efficientnet} & 12.0  & 1114 & \tzo1.8  &  4826  & 81.6  & 70.6\\
	 EfficientNet-B4~\cite{tan2019efficientnet} & 19.0  & \pzo573 & \tzo4.2 &  10006  & 82.9  & 72.3\\
	 EfficientNet-B5~\cite{tan2019efficientnet} & 30.0  & \pzo268 & \tzo9.9 &  11046  & 83.6  & 73.6\\
	 \midrule
	 NFNet-F0~\cite{Brock2021HighPerformanceLI} & 71.5 & \pzo950 & 12.4 & 4338 & 83.6  & 72.6 \\
	 NFNet-F1~\cite{Brock2021HighPerformanceLI} & 132.6 & \pzo337 & 35.5 & 6628 & 84.7  & 74.4\\
	 NFNet-F2~\cite{Brock2021HighPerformanceLI} & 193.8 & \pzo184 & 62.6 & 8144 & 85.1  & 74.3\\
	 NFNet-F3~\cite{Brock2021HighPerformanceLI} & 254.9 & \pzo101 & 115.0 & 11240 & 85.7  & 75.2 \\
	 NFNet-F4~\cite{Brock2021HighPerformanceLI} & 316.1 & \dzo59 & 215.3 & 16587 & 85.9  & 75.2 \\

\toprule
\multicolumn{7}{c}{\textbf{Vision Transformers and derivatives}} \\ [5pt]

    ViT: DeiT-T~\cite{Touvron2020TrainingDI}   & 5.7  & 3774 & \tzo1.3  &   536 & 72.2 &  60.4\\
    ViT: DeiT-S~\cite{Touvron2020TrainingDI,wightman2021resnet}  & 22.0  & 1891 & \tzo4.6 & 987 & 80.6 &  69.4\\
	ViT: DeiT-B~\cite{Touvron2020TrainingDI}    & 86.6  & \pzo831  & \dzo17.5 & 2078 & 81.8 &  71.5\\
	ViT: DeiT-B$\uparrow 384$~\cite{Touvron2020TrainingDI}   & 86.6  & \pzo195 & \dzo55.5 & 8956  & 83.1 & 72.4 \\
	\midrule
	Swin-T-224~\cite{liu2021swin} & 28.3 & 1109 & 4.5 & 3345 & 81.3 &  69.5 \\
    Swin-S-224~\cite{liu2021swin} & 49.6 & \pzo718 & 8.7 & 3470 &  83.0 &   71.8 \\

    Swin-B-224~\cite{liu2021swin} & 87.8  & \pzo532 & 15.4 & 4695 & 83.5 &   \_ \\
    Swin-B-384~\cite{liu2021swin} & 87.8  & \pzo159 & 47.0 & 19385 & 84.5 &   \_ \\
    \midrule
	CaiT-S24~\cite{touvron2021going}         & 46.9 & \pzo470 & 9.4 & 1469 & 82.7 &  \_ \\
	CaiT-M36~\cite{touvron2021going}         & 271.2  & \pzo159 & 53.7 & 3828 & 83.8  & \_\\

    \midrule
	
    XciT-S-12/16~\cite{el2021xcit} & 26.3 & 1372 & 4.8  & 1330 & 82.0   & \_ \\
    XciT-S-24/16~\cite{el2021xcit} & 47.7 & \pzo730 & 9.1  & 1452 & 82.6   & \_\\
    XciT-M-24/16~\cite{el2021xcit} & 84.4 & 545.8 & 16.2 & 2010.7 & 82.7 & \_  \\

    \toprule
\multicolumn{7}{c}{\textbf{Vision MLP}} \\[3pt]
    ResMLP-S12~\cite{Touvron2021ResMLPFN} &  15.0  & 3301 & \tzo3.0  & 755 &  76.6 &  64.4\\
    ResMLP-S24~\cite{Touvron2021ResMLPFN} &  30.0  &  1681    & \tzo6.0  & 844 &  79.4 &  67.9 \\
    ResMLP-B24~\cite{Touvron2021ResMLPFN} &  116.0 &      1120    & \dzo23.0 & 930  &   81.0 &  69.0 \\
    \toprule
    \multicolumn{7}{c}{\textbf{Patch-based ConvNets}} \\[3pt]
    ResMLP-S12 conv3x3~\cite{Touvron2021ResMLPFN} &  16.7  & 3217 & \tzo3.2  & 763 &  77.0  & 65.5\\
    ConvMixer-768/32~\cite{anonymous2022patches} & 21.1 & \pzo271 & 20.9 & 2644 & 80.2  & \_\\
    ConvMixer-1536/20~\cite{anonymous2022patches} & 51.6 & \pzo157 & 51.4 & 5312 & 81.4  & \_\\
    \midrule
    \rowcolor{Goldenrod}
    \ours-S36 & 16.2  & 1799  & 2.6 & 1270 &  80.7 &  69.7 \\
    \rowcolor{Goldenrod}
    \ours-S60 & 25.2  & 1125 & 4.0 & 1321 & 82.1 &  71.0\\
    \rowcolor{Goldenrod}
    \ours-S120&  47.7 & \pzo580 & 7.5 & 1450 & 83.2 &  72.5\\
    \rowcolor{Goldenrod}
    \ours-B60 &  99.4 & \pzo541 & 15.8 & 2790 & 83.5 &   72.6\\
    \rowcolor{Goldenrod}
    \ours-B120 & 188.6  & \pzo280 & 29.9  & 3314 & 84.1 & 73.9\\
    \midrule
    \rowcolor{Goldenrod}
    \ours-S60$\uparrow384$ & 25.2  & \pzo392 & 11.8 & 3600 & 83.7 &  73.4 \\
    \rowcolor{Goldenrod}
   \ours-B120$\uparrow384$& 188.6  & \dzo96 & 87.7 & 7587 & 85.2  & 75.6\\
    \bottomrule
    \end{tabular}}
\end{table}

\section{Transfer Learning experiments}
\label{sec:transfer}

We evaluate our architecture on 6 transfer learning tasks. The datasets used are summarized Table~\ref{tab:dataset}.  
For fine-tuning we used the procedure used in CaiT~\cite{touvron2021going} and DeiT~\cite{Touvron2020TrainingDI}.
Our results are summarized Table~\ref{tab:transfer}.
We can observe that our architecture achieves competitive performance on transfer learning tasks.

\begin{table}
\caption{Datasets used for our transfer learning tasks.  \label{tab:dataset}}
\centering
\scalebox{0.9}{
\begin{tabular}{l|rrr}
\toprule
Dataset & Train size & Test size & \#classes   \\
\midrule
iNaturalist 2018~\cite{Horn2018INaturalist}& 437,513   & 24,426 & 8,142 \\ 
iNaturalist 2019~\cite{Horn2019INaturalist}& 265,240   & 3,003  & 1,010  \\ 
Flowers-102~\cite{Nilsback08}& 2,040 & 6,149 & 102  \\ 
Stanford Cars~\cite{Cars2013}& 8,144 & 8,041 & 196  \\  
CIFAR-100~\cite{Krizhevsky2009LearningML}  & 50,000    & 10,000 & 100   \\ 
CIFAR-10~\cite{Krizhevsky2009LearningML}  & 50,000    & 10,000 & 10   \\ 
\bottomrule
\end{tabular}}
\end{table}

\begin{table}
    \caption{Results in transfer learning. 
    \label{tab:transfer}}
    \centering
    \scalebox{0.7}{
    \begin{tabular}{l|cccccc|r}
    \toprule
    Model                                      
        & \rotatebox{90}{CIFAR-10}
        & \rotatebox{90}{CIFAR-100}  
        & \rotatebox{90}{Flowers} 
        & \rotatebox{90}{Cars} 
        & \rotatebox{90}{iNat-18} 
        & \rotatebox{90}{iNat-19} 
        & \rotatebox{90}{FLOPs}\\
    \midrule

    ResNet-50~\cite{wightman2021resnet} & 98.3 & 86.9 & 97.9 & 92.7   & \_   & 73.9   &  4.1B\\
    Grafit~\cite{Touvron2020GrafitLF}  & \_& \_ & 98.2 & 92.5   & 69.8  & 75.9   & 4.1B\\
    \midrule
    EfficientNet-B7~\cite{tan2019efficientnet}  & 98.9 & 91.7  & 98.8 & 94.7 & \_ & \_ & 37.0B \\
    \midrule       
    ViT-B/16~\cite{dosovitskiy2020image} & 98.1 & 87.1 & 89.5 & \_   & \_   & \_   &  55.5B\\
    ViT-L/16~\cite{dosovitskiy2020image}& 97.9 & 86.4 & 89.7 & \_   & \_   & \_   &  190.7B\\
    DeiT-B~\cite{Touvron2020TrainingDI} & 99.1 & 90.8 & 98.4 & 92.1 & 73.2 & 77.7 &  17.5B \\
    CaiT-S-36~\cite{touvron2021going}   & 99.2 & 92.2 & 98.8 & 93.5 & 77.1 & 80.6 & 13.9B\\
    CaiT-M-36~\cite{touvron2021going}   & \textbf{99.3} & \textbf{93.3} & 99.0 & 93.5 & 76.9 & 81.7 & 53.7B\\
    \midrule   
    Ours-S60  & 99.2  & 91.4 & 98.8  & 94.0 & 72.9 & 78.1 & 4.0B \\
    Ours-B120 & 99.2  & 91.1  & 99.0 & 94.4 & 74.3 & 79.5 & 29.9B \\
    \midrule   
    Ours-S60 @ 320 & 99.1  & 91.4 & 98.9 & 94.5 & 76.8 & 81.4 & 8.2B \\
    Ours-B120 @ 320 &99.1  & 91.2 & \textbf{99.1} & \textbf{94.8} & \textbf{79.6} & \textbf{82.5} & 60.9B \\
    \bottomrule
    \end{tabular}}
\end{table}

%% file: supplementary_title.tex
\appendix

\clearpage
\counterwithin{figure}{section}
\counterwithin{table}{section}
\counterwithin{equation}{section}

\pagenumbering{Roman}  
\pagestyle{plain}

\twocolumn[
\newpage
\null
\vskip .375in
\begin{center}

{\Large \bf \inserttitle ~ \\ \vspace{0.5cm} \large Supplementary Material \par}
  \vspace*{24pt}
  {\par}
\end{center}]